\documentclass{egpubl}
\usepackage{eg2026}
 
\ConferencePaper         
\CGFccby

\usepackage[T1]{fontenc}
\usepackage{dfadobe}  

\biberVersion
\BibtexOrBiblatex
\usepackage[backend=biber,bibstyle=EG,citestyle=alphabetic,backref=true]{biblatex} 
\electronicVersion
\PrintedOrElectronic

\ifpdf \usepackage[pdftex]{graphicx} \pdfcompresslevel=9
\else \usepackage[dvips]{graphicx} \fi

\usepackage{egweblnk} 
\author[D. Lu et al.]
{
\parbox{\textwidth}{\centering
Dongyue Lu\footnotemark[1]$^{1}$,
Ao Liang\footnotemark[1]$^{1}$,
Tianxin Huang$^{2}$,
Xiao Fu$^{3}$,
Yuyang Zhao$^{1}$,
Baorui Ma$^{4}$,
Liang Pan$^{5}$,
Wei Yin$^{6}$,
Lingdong Kong\footnotemark[2]$^{1}$,
Wei Tsang Ooi$^{1}$,
and Ziwei Liu$^{7}$
}
\\
\parbox{\textwidth}{\centering
$^{1}$NUS,
$^{2}$HKU,
$^{3}$CUHK,
$^{4}$THU,
$^{5}$Shanghai AI Lab,
$^{6}$Horizon Robotics,
$^{7}$NTU
}
}

\def\etc{\textit{etc.}}
\def\ie{\textit{i.e.}}

\usepackage{booktabs}
\usepackage{amsmath, amssymb}
\usepackage{graphicx}
\usepackage{hyperref}
\usepackage{cleveref}
\usepackage{enumitem}
\usepackage{fontawesome}
\usepackage{multirow}
\usepackage[table,dvipsnames]{xcolor}

\definecolor{see_purple}{RGB}{166,82,166}
\definecolor{see_green}{RGB}{0,176,80}
\definecolor{see_yellow}{RGB}{229,157,35}
\definecolor{see_gray}{RGB}{165,165,165}
\definecolor{link}{RGB}{166,82,166}

\definecolor{see_1}{RGB}{67,89,216}
\definecolor{see_2}{RGB}{95,90,222}
\definecolor{see_3}{RGB}{124,91,228}
\definecolor{see_4}{RGB}{151,92,234}
\definecolor{see_5}{RGB}{180,93,240}

\definecolor{link}{RGB}{166,82,166}
\definecolor{rev}{RGB}{0,0,0}
\definecolor{refine}{RGB}{0,0,0}
\hypersetup{
  colorlinks=true,
  linkcolor=link,
  citecolor=link,
  urlcolor=link,
}

\newcommand{\ours}{\textsc{\textcolor{see_1}{S}\textcolor{see_2}{e}\textcolor{see_3}{e}\textcolor{see_4}{4}\textcolor{see_5}{D}}}
\newcommand{\rev}[1]{\textcolor{rev}{#1}}
\newcommand{\refine}[1]{\textcolor{refine}{#1}}

\title[\ours]
{\scalebox{0.98}{\ours: Pose-Free 4D Generation via Auto-Regressive Video Inpainting}}
\begin{document}

\teaser{
 \vspace{-1.2cm}
 \includegraphics[width=\linewidth]{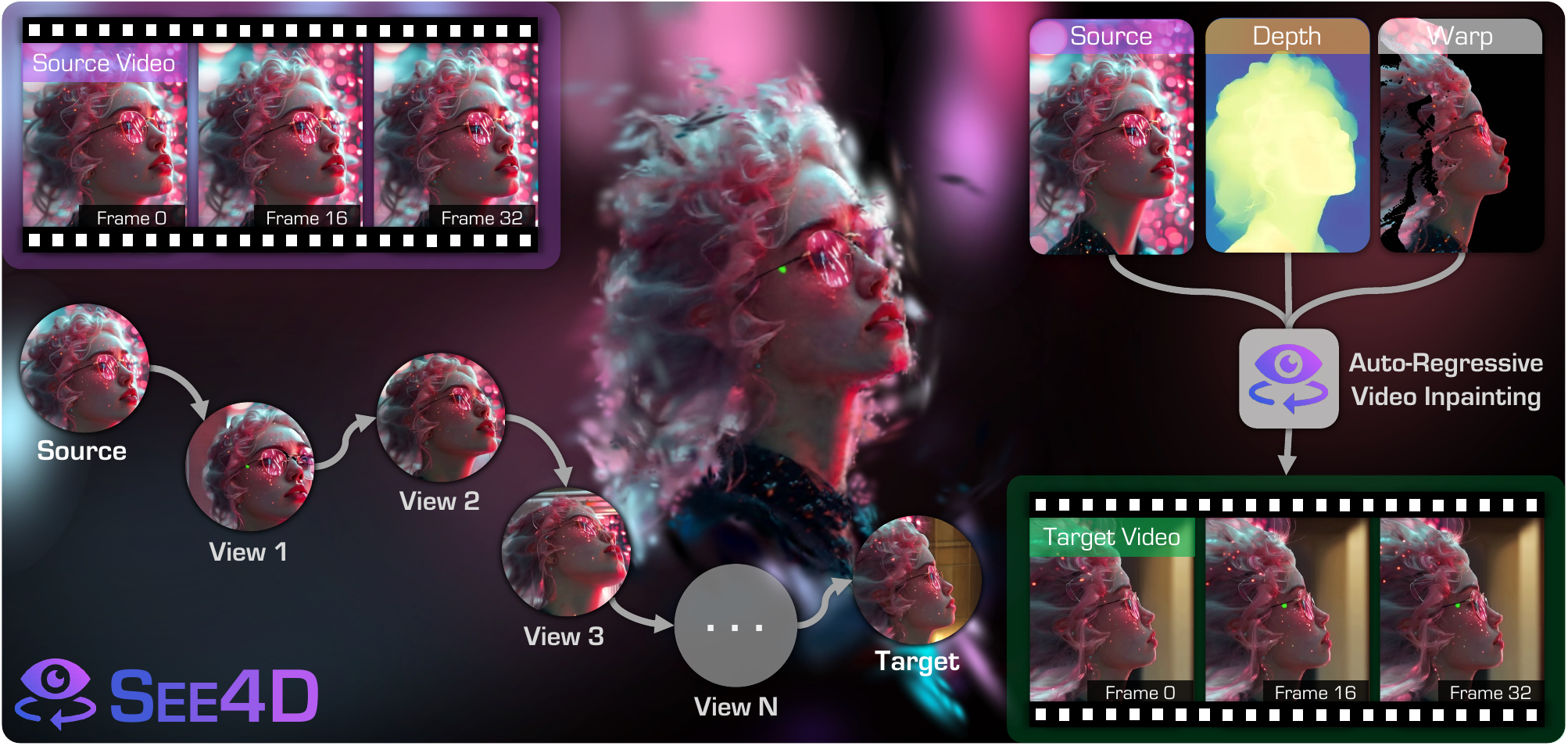}
 \centering
 \vspace{-0.5cm}
\caption{\textbf{Illustration of our 4D generation framework} (\ours). 
\refine{Given an unposed source video, we spline-interpolate a virtual camera trajectory and estimate per-frame depth to lift frames into 3D. Depth-guided forward warping yields intermediate latents, which, together with source latents, are processed by a view-conditional diffusion model with spatiotemporal attention and noise-adaptive conditioning in an autoregressive manner to synthesize the target-view sequence. Additional demos are provided in the supplementary materials.}
}
\label{fig:teaser}
}

\maketitle
\begin{abstract}
Immersive applications call for synthesizing spatiotemporal 4D content from casual videos without costly 3D supervision. Existing video-to-4D methods typically rely on manually annotated camera poses, which are labor-intensive and brittle for in-the-wild footage. Recent warp-then-inpaint approaches mitigate the need for pose labels by warping input frames along a novel camera trajectory and using an inpainting model to fill missing regions, thereby depicting the 4D scene from diverse viewpoints. However, this trajectory-to-trajectory formulation often entangles camera motion with scene dynamics and complicates both modeling and inference. We introduce \ours, a pose-free, trajectory-to-camera framework that replaces explicit trajectory prediction with rendering to a bank of fixed virtual cameras, thereby separating camera control from scene modeling. A view-conditional video inpainting model is trained to learn a robust geometry prior by denoising realistically synthesized warped images and to inpaint occluded or missing regions across virtual viewpoints, eliminating the need for explicit 3D annotations. Building on this inpainting core, we design a spatiotemporal autoregressive inference pipeline that traverses virtual-camera splines and extends videos with overlapping windows, enabling coherent generation at bounded per-step complexity. We validate See4D on cross-view video generation and sparse reconstruction benchmarks. Across quantitative metrics and qualitative assessments, our method achieves superior generalization and improved performance relative to pose- or trajectory-conditioned baselines, advancing practical 4D world modeling from casual videos.

\ccsdesc[300]{Computing methodologies~Reconstruction}
\ccsdesc[300]{Computing methodologies~Appearance and texture representations}

\printccsdesc   
\end{abstract}  

\section{Introduction}
\label{sec:intro}

Virtual Reality (VR) has evolved from a niche curiosity into a pervasive medium that impacts storytelling, scientific communication, and professional training~\cite{burdea2003virtual, zheng1998virtual,wohlgenannt2020virtual}. A truly immersive experience, however, demands more than static panoramic images. It requires the ability to wander freely through a scene as it unfolds over time, i.e., a 4D environment that is simultaneously spatially and temporally consistent. Delivering such content calls for densely synchronized, multi-view videos captured with tightly calibrated camera arrays~\cite{lavalle2023virtual, liang2025worldlens, kong20253d, anthes2016state}. These systems are costly, fragile, and generally impractical outside controlled studios, forcing most consumer VR experiences to rely on sparse or synthetic assets. Bridging this gap by converting a single, casually captured video into a coherent 4D representation would unlock an enormous amount of real-world footage for VR content creation, motivating the recent surge of research in \textbf{video-to-4D} generation.

Current efforts address this problem along two principal directions. 
The first class of methods conditions the model on \textbf{explicit poses}, assuming that each frame is paired with a precise external pose, 
and trains the network to generate novel views under these specified poses~\cite{bai2024syncammaster,bai2025recammaster,zhao2025genxd,wu2024cat4d, yu2024viewcrafter}.
\rev{While effective in simulation, these methods rely on explicit pose annotations defined in a specific coordinate system and require the model to directly condition on pose representations during training,}
which are difficult to obtain in the dynamic, hand-held videos common on online platforms.
The second class leverages monocular depth estimation and employs a \textbf{warp-then-inpaint} strategy to synthesize novel views~\cite{ma2025see3d,yu2025trajectorycrafter,huang2025vivid4d, ren2025gen3c,gu2025diffusion}. 
Pixels are first projected into the target view using estimated depth~\cite{yang2024depth, hu2024depthcrafter}, yielding an incomplete approximation constrained by visible geometry. 
An inpainting model, conditioned on the \textbf{warped image}, is then employed to hallucinate the missing content. 
Compared to pose-based conditioning, warped-image conditioning generally achieves stronger 3D consistency by explicitly encoding geometric cues and preserving cross-view coherence.

Despite this superiority, most warp-then-inpaint systems are optimized for \textbf{trajectory-to-trajectory} synthesis: given a video captured along a specific camera path, they synthesize a novel video from an entirely different trajectory, preserving synchronized scene dynamics~\cite{yu2025trajectorycrafter, bai2024syncammaster, bai2025recammaster, hoorick2024gcd, huang2025vivid4d, ren2025gen3c}. 
This dual-trajectory formulation, however, is relatively under-constrained, which can complicate training and lead to less stable inference. The generated viewpoints often remain close to the input path, and the reconstruction quality can be sensitive to per-frame depth accuracy. Consequently, these approaches face challenges in scaling to full 4D scene reconstruction that enables flexible free-viewpoint playback, a capability that is particularly important for immersive VR experiences.

To address this challenge, we propose \ours, a pose-free framework based on a \textbf{trajectory-to-camera} formulation. Our method generates a bank of synchronized, fixed-view videos from a single monocular clip. 
\rev{We train a view-conditional inpainting model using depth-warped images and their associated masks, rather than relying on explicit camera poses.
By providing implicit geometric guidance through warped observations instead of explicit camera pose matrices, this design enables pose-free novel-view conditioning, while reducing sensitivity to pose errors and dependence on specific coordinate conventions.}
To ensure cross-view and cross-frame coherence, we integrate a lightweight spatial–temporal transformer backbone that jointly models spatial and temporal consistency.
During training, we apply \textit{realistic warp synthesis} by forward-projecting the target-view video to a nearby random pose and back-projecting with jitter, simulating the source-to-target warp artifacts observed during inference.
Additionally, we propose \textit{noise-adaptive condition}, which adds modulated noise to the conditional warped images based on their corresponding warp mask, preventing overfitting to unreliable warps.
This design simplifies optimization compared to \textbf{trajectory-to-trajectory} methods, remains robust in dynamic and complex scenarios, and produces high-quality, geometrically consistent novel views suitable for large-scale 4D scene modeling and representation.

Complementing the core model, we introduce a \textbf{spatio-temporal auto-regressive} inference pipeline that robustly scales synthesis to long sequences and viewpoint shifts. Spatially, we march smoothly along a spline of virtual camera poses between the source and each target view, iteratively warping, inpainting, and updating depth to progressively mitigate cross-view self-occlusions. Temporally, we extend clips by overlapping successive diffusion runs, reusing a sliding window of one prediction as the seed for the next. This dual recursion effectively enforces smooth motion continuity while gracefully handling depth noise and complex disocclusion.

We evaluate our method on both 4D scene reconstruction accuracy and perceptual video quality. On public 4D reconstruction benchmarks, our method delivers the highest completeness and geometric fidelity, surpassing all previous methods. On our newly collected set of challenging hand-held videos, it also wins across standard generation metrics, achieving sharper detail, stable cross-view consistency, and stronger temporal coherence than prior work. 
Beyond numbers, we demonstrate practical benefits in robot manipulation, autonomous driving, interactive gaming, and movie production, underlining the real-world value of our pipeline.

The key contributions of this work can be summarized as:
\begin{itemize}
\item We propose \ours, a pose-free pipeline that learn to generate 4D scenes directly from in-the-wild videos, eliminating the need for large-scale pose annotations.
\item We propose a trajectory-to-camera formulation. By predicting a set of fixed target views rather than a full camera path, we simplify training and improve synthesis stability.
\item We introduce a spatial-temporal auto-regressive inference process that delivers long, temporally consistent videos while coping with self-occlusion and depth noise.
\item Extensive evaluations show that our method outperforms existing baselines in both reconstruction accuracy and generative quality, unlocking new workflows for VR content creation.
\end{itemize}
\section{Related Work}
\label{sec:related_work}

\subsection{3D \& 4D Reconstruction}
Early neural rendering methods reconstruct static scenes from sparse multi-view supervision or single-image priors using NeRF-style implicit fields~\cite{wang2021nerf,barron2021mipnerf,barron2022mipnerf360,barron2023zipnerf,miao2024edus,haque2023in2n,irshad2022centersnap,knapitsch2017tnt,sargent2023zeronvs} and more recently, Gaussian-splatting variants~\cite{kerbl2023-3dgs,zhou2024hugs,yan2024streetgs,huang2024-2dgs,yu2024mipsplatting,guedon2024sugar,charatan2024pixelsplat,lu2024scaffoldgs}. To reduce ambiguity from limited views, follow-up work incorporates classical geometry and learned priors by bootstrapping depth, flow, or correspondences~\cite{fan2024instantsplat,yang2023freenerf,xin2023simplemapping, chen2024mvsplat,zhang2024corgs}, or regularizing radiance fields with geometric constraints~\cite{yang2023freenerf,pumarola2021dnerf,niemeyer2022regnerf,wang2023sparsenerf,li2024dngaussian}. Others accelerate inference via feed-forward prediction of point clouds, meshes, or Gaussians~\cite{szymanowicz2025bolt3d,wang2023dust3r}, trading some fidelity for efficiency. Extending to 4D, NeRF-based models~\cite{pumarola2021dnerf,yoon2020novel,du2021neural,li2021neural,li2023dynibar,park2021nerfies,park2021hypernerf,wang2021neural,xian2021space} and dynamic Gaussians~\cite{wu20244d,duan20244d,yang2023real,yang2024deformable,iphone} handle moderate motion with multi-view videos. Monocular scenarios are more challenging; for instance, Shape-of-Motion~\cite{wang2024shape} fuses depth, tracking, and masks, yet struggles with occlusion. Recent methods synthesize pseudo-multi-view observations from monocular frames to support downstream reconstruction~\cite{zhao2025genxd,huang2025vivid4d}. Nevertheless, achieving coherent and practical 4D reconstruction remains an open problem.

\subsection{Generative Novel View Synthesis}
Diffusion models now serve as flexible priors for integrating geometry in various ways. Score-distillation pipelines optimize 3D representations to match image-level diffusion gradients~\cite{zhou2023sparsefusion,chen2024svsgs,poole2022dreamfusion}, while others inject structural cues like depth into diffusion backbones, enhancing fidelity but depending on static content and accurate poses~\cite{wu2024reconfusion,ma2025see3d,yu2024viewcrafter,yu2024lmgs,liu2024viewextrapolator,seo2024genwarp,liu2024reconx,gao2024cat3d,wang2023dust3r,wang2021nerf,kerbl2023-3dgs,zhao2025genxd}. Extending these to 4D requires modeling motion and scene change, pose-aware or short-range approaches~\cite{zhao2025genxd,liu2023zero,wu2024cat4d,park2025steerx} often struggle with complex dynamics, as shown by WideRange4D~\cite{yang2025widerange4d}. Other directions refine trajectories~\cite{ji2025posetraj} or distill geometry from pretrained models~\cite{liu2025free4d}, yet large-scale consistency remains challenging. Meanwhile, video inpainting showcases diffusion’s temporal coherence~\cite{zi2024cococo,li2025diffueraser,gu2024floed,zhang2024avid,lee2024fffvdi,fan2023m3ddm,wang2024motia,chen2024follow}, motivating warp-then-inpaint pipelines. See3D~\cite{ma2025see3d} learns implicit geometry from pose-free videos, and BlobCtrl~\cite{li2025blobctrl} links 2D generation with 3D edits via Gaussians. Broader view synthesis uses masked views from stereo or synthetic geometry~\cite{zhao2024stereocrafter,bian2025gs-dit} and depth-informed masks for fine-tuned diffusion models~\cite{ren2025gen3c,yu2025trajectorycrafter,park2025zero4d,yao2025sv4d-2,blattmann2023svd,hong2022cogvideo,yang2024cogvideox}. Though effective, current warp-then-inpaint pipelines demand heavy preprocessing and degrade on long videos. Achieving spatial and temporal generality without curated multi-view input remains unsolved, which is a gap our pose-free, autoregressive approach aims to address.

\subsection{Camera-Controlled Video Generation}
Recent progress in video diffusion has enabled camera-controlled generation, where models follow user-defined viewpoint paths~\cite{bai2025recammaster, he2024cameractrl, he2025cameractrl-2, jeong2025reangle, bai2024syncammaster, hoorick2024gcd}. Early methods embed 6DoF poses into 2D or video diffusion backbones~\cite{zhao2025genxd, ren2025gen3c, bai2025recammaster}, while TrajectoryCrafter~\cite{yu2025trajectorycrafter} fine-tunes video models on multi-view warped data to align with target geometry. CameraCtrl~\cite{he2024cameractrl} adopts classifier-free guidance to support diverse trajectories but struggles with large viewpoint shifts. CameraCtrl II~\cite{he2025cameractrl-2} improves robustness by adding new training data and a pose-injection module, enabling long-range exploration and consistent scene stitching. Reangle-A-Video~\cite{jeong2025reangle} uses a video-to-video framework that fine-tunes on warped frames, generating smooth cross-view transitions. GS-DiT~\cite{bian2025gs-dit} and StereoCrafter~\cite{zhao2024stereocrafter} combine warped views with diffusion priors, but depend on synthetic training and are limited to moderate motion. Together, these works pave the way for interactive 4D generation, where users freely control camera movement and witness scenes unfold from novel perspectives.
\section{Methodology}
\label{sec:method}

We introduce \ours, a pose-free trajectory-to-camera framework that synthesizes consistent 4D scenes from a single monocular video. Unlike trajectory-to-trajectory methods, it renders to a set of fixed virtual cameras, thereby decoupling scene modeling from camera motion and improving stability. \Cref{sec:preliminaries} reviews the foundations of latent diffusion and image warping. \Cref{sec:inpaint_model} presents our view-conditional inpainting model. \Cref{sec:inference} describes our auto-regressive inference pipeline that maintains spatial-temporal consistency.

\subsection{Preliminaries}
\label{sec:preliminaries}

\begin{figure*}[t]
    \centering
    \includegraphics[width=\linewidth]{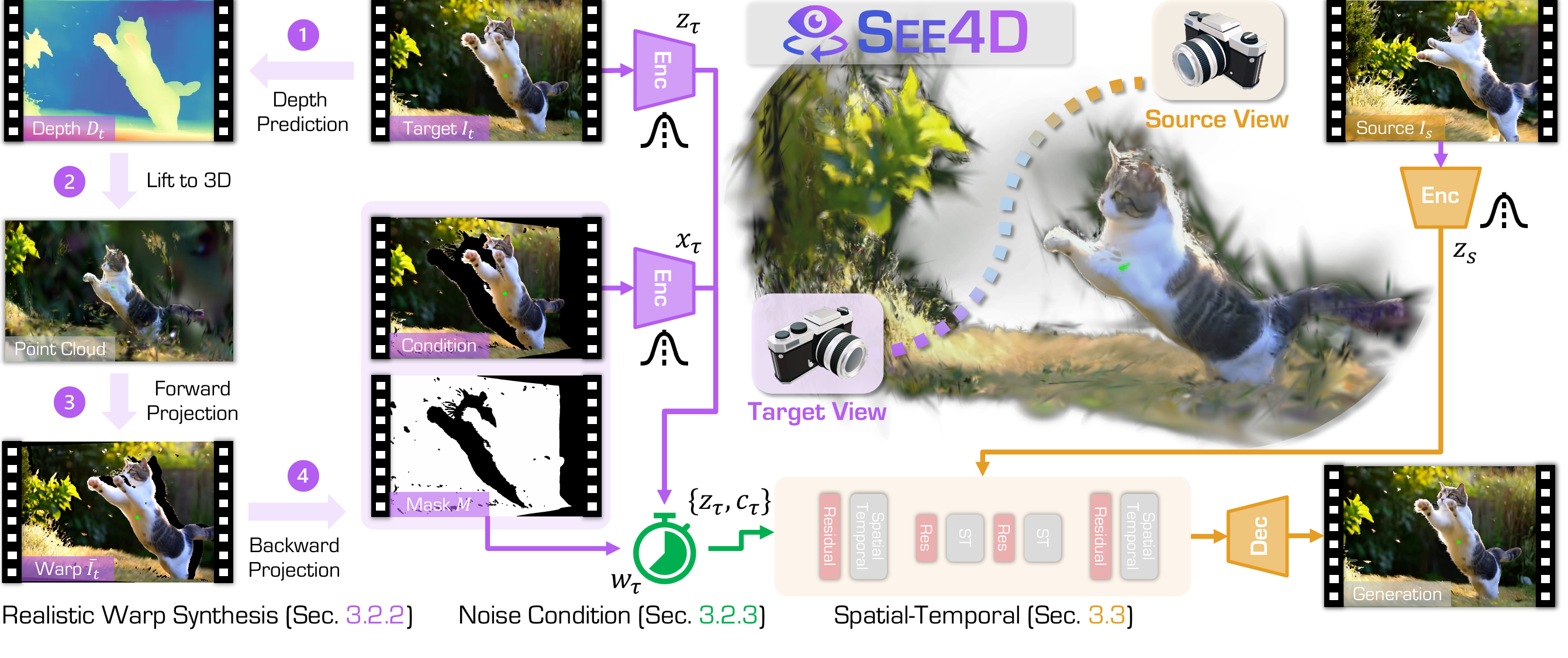}
    \caption{\textbf{Overview of our view-conditional inpainting model}. From a synchronized multi-view dynamic scenario, we select a source–target view pair. We apply realistic warp synthesis (\textit{cf}.~Sec.\ref{sec:Realistic Warp Synthesis}) to generate a warp image and mask as conditioning. The warp latent is then perturbed in a noise-adaptive manner (\textit{cf}.~Sec.\ref{sec:Noise‑Adaptive Condition}). Finally, the source latent, warp condition, and noisy target latent are input to our spatial-transformer backbone (\textit{cf}.~Sec.\ref{sec:Spatial–Temporal Backbone}) to denoise the latent and inpaint the masked regions. These components yield a pose-free inpainting model stable on dynamic and complex footage.}
    \label{fig:overview}
    \vspace{-0.5cm}
\end{figure*}

\subsubsection{Latent Diffusion Model}
We adopt a latent diffusion model (LDM) \cite{rombach2022high} as the generative backbone for our fixed‐view synthesis. We first compress an input video $x$ into a latent tensor $z_{0}=\mathcal{E}(x)$ using a pre-trained VAE \cite{kingma2013auto} encoder $\mathcal{E}$.  
A variance-preserving forward process adds Gaussian noise, \ie,
\begin{equation}
    z_{\tau}=\sqrt{\bar\alpha_{\tau}}\,z_{0}+\sqrt{1-\bar\alpha_{\tau}}\,\varepsilon,\qquad
    \varepsilon\!\sim\!\mathcal{N}(0,I),
\end{equation}
where $\bar\alpha_{\tau}$ is a fixed scheduler and $\tau\!\in\![1,T]$.  
A denoiser $\varepsilon_{\theta}$ is trained to predict $\varepsilon$ by minimizing the following term:
\begin{equation}
    \mathcal{L}_{\mathrm{LDM}}=
    \mathbb{E}_{z_{0},\varepsilon,\tau}
    \bigl[\,
        \lVert\varepsilon_{\theta}(z_{\tau},c,\tau)-\varepsilon\rVert_{2}^{2}
    \bigr],
\end{equation}
where $c$ denotes the controllable condition. At test time, iterative denoiser transforms $z_{T}\!\sim\!\mathcal{N}(0,I)$ back to $z_{0}$, which the decoder maps to the output video.  
Working in latent space cuts computation by an order of magnitude while preserving perceptual quality.

\subsubsection{Depth-Guided Image Warp}
We warp source frames using predicted depth to build a view-specific prior without explicit poses.  
Given a per–pixel depth map $\mathbf{D}$, camera intrinsics $\mathbf{K}$, and a relative transform $\mathbf{T}\in\mathrm{SE}(3)$, each pixel $\mathbf{u}=(u,v)$ is lifted to 3D via homogeneous back-projection, mapped to the target frame, and then re-projected. The process is formulated as:
\begin{equation}
    \mathbf{u}_w=
    \pi\bigl(
        \mathbf{K}\mathbf{T}
        \bigl[\mathbf{D}(\mathbf{u})\mathbf{K}^{-1}[u,v,1]^{T};1\bigr]
    \bigr),
\label{eq:warp}
\end{equation}
where $\pi([u,v,w]^{T})=[u/w,\;v/w]^{T}$. We bilinearly splat the source pixels to sub-pixel targets, producing a warped frame $\mathbf{I}_w$ and a mask $\mathbf{M}$ that flags depth clashes or off-screen projections. This sparse yet structured pair conditions the inpainting diffuser, enforcing target-view geometry while letting it hallucinate missing regions.

\subsection{View‑Conditional Inpainting Model}
\label{sec:inpaint_model}

\subsubsection{Overview} Obtaining frame-accurate 6-DoF poses for in-the-wild videos is costly and unreliable. Instead, we adopt the warp-then-inpaint paradigm \cite{ma2025see3d,yu2025trajectorycrafter,ren2025gen3c}. During training, given paired source and target videos $(\mathbf{I}_{s},\mathbf{I}_{t})$, for each target frame $\mathbf{I}_{t}$, we estimate its depth and warp it to a virtual pose using Eq.~\eqref{eq:warp}. This produces a pixel-aligned warped image $\mathbf{I}_{w}$ together with a visibility mask $\mathbf{M}$. We then construct the conditioning tensor $c_{\tau}$ by fusing $\mathbf{I}_{w}$, $\mathbf{M}$, and the step-dependent noisy latent $z_{\tau}$. The denoiser is trained with the objective
\begin{equation}
    \mathcal{L}=
    \mathbb{E}_{z_{0},z_s,\varepsilon,\tau}
    \bigl[\,
        \lVert\varepsilon_{\theta}(z_{\tau},z_s,c_{\tau},\tau)-\varepsilon\rVert_{2}^{2}
    \bigr],
\label{eq:loss}
\end{equation}
where $z_s$ is the clean latent of the source video.

Prior warp{\mbox-}conditioned models~\cite{ma2025see3d,yu2025trajectorycrafter,huang2025vivid4d} are tuned for static scenes but fail under fast motion or viewpoint variations. We address this with: 
\textbf{(1) Realistic Warp Synthesis.} 
To simulate real-world warp artifacts observed during source-to-target projection at inference time, we perturb the target view by forward-projecting it to a randomly sampled nearby virtual camera pose, followed by back-projection using the jittered pose during training.
\textbf{(2) Noise-Adaptive Condition.} 
We introduce adaptive noise to the conditional warped images based on their corresponding warp masks, helping prevent the model from overfitting to unreliable warping during training.
\textbf{(3) Spatial-Temporal Backbone.} 
We introduce a lightweight spatio-temporal transformer to enforce cross-view and cross-frame coherence by incorporating frame-time embeddings and applying spatio-temporal attention.
Together, these components form a view-conditional inpainting model that maintains stability in dynamic scenes while preserving the simplicity of pose-free supervision.

\subsubsection{Realistic Warp Synthesis} 
\label{sec:Realistic Warp Synthesis}
Existing methods conditioned on warped images targets with random blocks \cite{ma2025see3d}, use 2D trackers for motion estimation \cite{huang2025vivid4d}, or ignore depth noise artifacts \cite{yu2025trajectorycrafter}. These heuristics miss two key artifacts in dynamic video: \textit{camera motion} that pushes content off the frame and \textit{object motion} that causes irregular object gaps. In addition, noisy monocular depths also add tearing patterns that simple masks cannot capture.

Existing methods conditioned on warped images targets with random blocks \cite{ma2025see3d}, use 2D trackers for motion estimation \cite{huang2025vivid4d}, 
or ignore depth noise artifacts \cite{yu2025trajectorycrafter}. 
These heuristics miss two key artifacts in dynamic video: \textit{camera motion} that pushes content off the frame and \textit{object motion} that causes irregular object gaps. In addition, noisy monocular depths also add tearing patterns that simple masks cannot capture. \rev{Notably, while TrajectoryCrafter \cite{yu2025trajectorycrafter} employs a double re-projection strategy to simulate geometric inconsistencies, it does not explicitly account for foreground-driven motion patterns or depth-induced distortions prevalent in dynamic videos.}

A perfect supervision signal would attach scale-aligned depths and ground-truth poses to every source-target video pair $(\mathbf{I}_{s},\mathbf{I}_{t}) $, but collecting dense 6-DoF labels is prohibitively expensive. Instead, for each target frame, we procedurally create a warped image \(\mathbf{I}_{w}\) and mask \(\mathbf{M}\) that imitate test-time noise through four steps: \textit{(1) Depth prediction.} We use a monocular estimator
\cite{yang2024depth} to produce \(\mathbf{D}_{t}\) and lift
\(\mathbf{I}_{t}\) into a point cloud. \textit{(2) Random scene re-pose.} We segment the foreground objects using \cite{meyer2025ben}, designate the point corresponding to the center of the largest inscribed circle of the largest object as the world origin, then rotate the point cloud about a random axis by \([-30^{\circ},\,30^{\circ}]\), then translate by 
$\mathbf{n}\sim\mathcal{N}\!\bigl(\mathbf{0},(\lambda\bar{D})^{2}\mathbf{I}_{3}\bigr)$ (\(\bar{D}\) is mean depth, \(\lambda\) is a scale factor, and $\mathbf{I}_{3}$ is the $3\times3$ identity matrix), yielding a synthetic transform \({\mathbf{T}_w}\). \rev{This foreground-centric rotation produces realistic disocclusions around moving objects, whereas arbitrary global rotations tend to introduce largely unstructured noise.} \textit{(3) Forward projection.} We render the transformed point cloud with Z-buffer and bilinear splatting to preserve correct visibility, obtaining a warped image \(\bar{\mathbf{I}}_{t}\) and depth \(\bar{\mathbf{D}}_{t}\), this introduces tearing and stretching artifacts characteristic of real warps. \textit{(4) Back-projection with noise.} We warp reversely with \(\bar{\mathbf{I}}_{t}\) and  \(\bar{\mathbf{D}}_{t}\). As monocular depths are noisy, we perturb the inverse transform \({\mathbf{T}_w^{-1}}\) with pose jitter $\delta\mathbf{T}$ to produce a slightly misaligned yet informative warp \(\mathbf{I}_w\) that mirrors test-time noise.  \rev{This noise injection explicitly mimics depth-induced distortions absent in prior double re-projection schemes.} The same Z-buffer yields the binary mask \(\mathbf{M}\), whose ragged contours precisely delineate disocclusions at borders and around moving objects, which is far richer than that in earlier work.

\subsubsection{Noise‑Adaptive Condition}
\label{sec:Noise‑Adaptive Condition}
Warp-conditioned diffusion models limit signal leakage by injecting step-dependent noise. Excessive noise degrades camera control, whereas insufficient noise encourages the network to replicate the warp. Dynamic inputs intensify this trade-off because depth errors are larger and data diversity is limited.  To curb over-reliance on the warp latent, we apply a stronger perturbation.
\begin{equation}
    x_{\tau}=\sqrt{\bar\alpha_{\tau/3}}\,z_{w}+
         \sqrt{1-\bar\alpha_{\tau/3}}\,\varepsilon,\quad
\varepsilon\sim\mathcal{N}(0,I),
\end{equation}
where $z_{w}$ is the latent encoding of the warped image, and $x_{\tau}$ denotes its noisy counterpart, the reduced index $\tau/3$ yields higher variance than that used for the target latent at the same step.

The informativeness of the warp condition directly impacts inpainting quality, unreliable warps should be down-weighted, while accurate ones should be preserved. We approximate warp fidelity by the mask density $\beta=\|\mathbf{M}\|_{0}/(BNHW)$, where $\|\mathbf{M}\|_{0}$ counts non-zero entries, $B$ is the batch size, $N$ the number of frames, and $H{\times}W$ the spatial resolution. A larger $\beta$ indicates greater pixel overlap and thus higher reliability. We therefore modulate the blend between $x_{\tau}$ and the running target latent $z_{\tau}$ using the mask density
\begin{equation}
    c_{\tau}=\bigl[\gamma\beta\,w_{\tau}x_{\tau}+(1-\gamma\beta\,w_{\tau})z_{\tau};\,\mathbf{M}\bigr],
\end{equation}
where $w_{\tau}\!\in\![0,1]$ decays monotonically with~$\tau$, and $\gamma$ is a scale factor. Hence, reliable warps transmit more signal, whereas sparse or noisy warps are increasingly suppressed. Guided by the time-dependent weight $w_{\tau}$, early iterations lock coarse geometry from $x_{\tau}$, while later iterations refine details through $z_{\tau}$, yielding effective error correction without sacrificing camera control.

\subsubsection{Spatial–Temporal Backbone}  
\label{sec:Spatial–Temporal Backbone}

Our architecture builds on the multi-view UNet of previous work \cite{ma2025see3d} and introduces two compact temporal mechanisms. \textit{(1) Frame-time embedding.} A sinusoidal encoding of the frame index is injected into every residual path, giving the feature hierarchy an explicit notion of temporal order. \textit{(2) Spatial–temporal attention.} Each 2D transformer block is upgraded to a module that attends simultaneously to spatial tokens and their counterparts in neighboring frames, enforcing cross-frame coherence. These additions provide the temporal context needed for consistent geometry and appearance throughout the sequence.

Our view-conditional inpainting network eliminates the need for per-frame pose annotations by conditioning on depth-warped images, thereby leveraging rich appearance cues without explicit calibration. Coupled with realistic warp synthesis, noise-adaptive conditioning, and the proposed spatial–temporal backbone, this design yields a lightweight yet temporally consistent model that generates high-quality novel views from pose-free, dynamic video clips.

\begin{figure}[t]
    \centering
    \includegraphics[width=\linewidth]{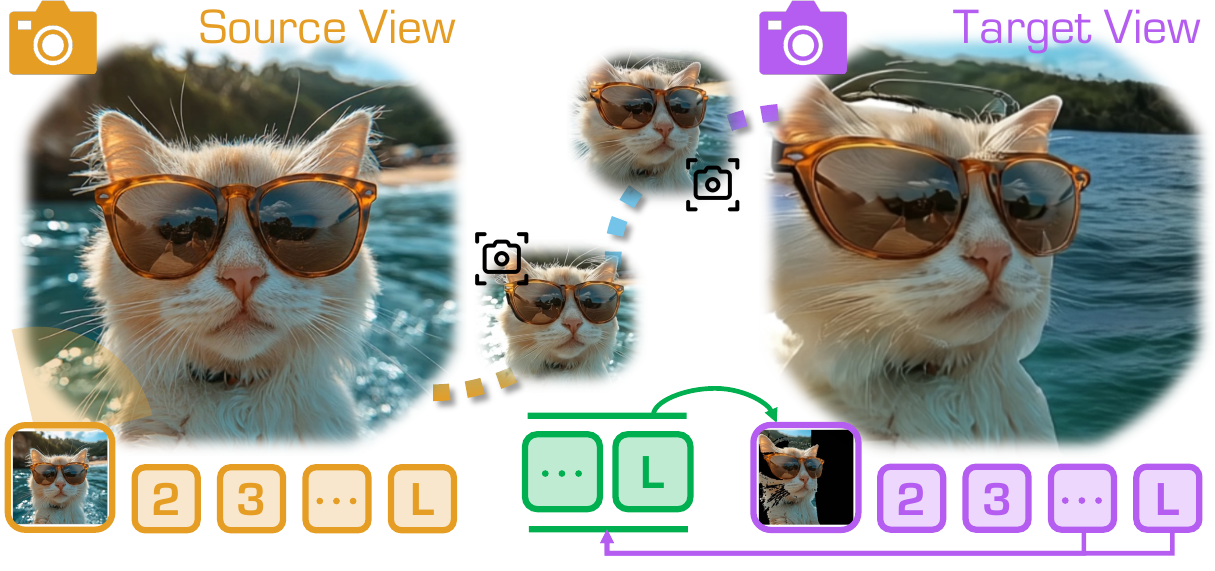}
    \caption{
    \textbf{Overview of our spatiotemporal auto‐regressive inference pipeline}. We decompose a single warp into small spline‐interpolated hops. We then prepend the denoised clean latents to subsequent noisy windows, producing seamless long videos without extra memory overhead.
    }
    \vspace{-0.6cm}
    \label{fig:inference}
\end{figure}

\subsection{Spatial-Temporal Auto-Regressive Inference}
\label{sec:inference}

At inference time, our system needs to convert a monocular dynamic sequence captured at a single view into a set of temporally aligned videos at multiple, user-specified fixed target viewpoints. These synthesized clips are later fused by downstream 4D pipelines~\cite{wu20244d} into a dense space-time representation. This task is limited by two factors. \textit{(1) Viewpoint shifts.} Even moderate angular gaps can significantly amplify depth inaccuracies, directly warping from the source pose to a distant camera often produces tearing and occlusion holes, resulting in an unreliable conditioning signal for the inpainting stage. \textit{(2) Long durations.} The diffusion backbone can denoise only a few dozen frames in a single forward pass, so naïvely processing a minute-long video breaks temporal coherence at window boundaries. We tackle both issues with an auto-regressive inference pipeline that proceeds \textbf{progressively in space}, advancing smoothly from the source view to the target view through a chain of virtual cameras, and \textbf{progressively in time}, sliding a partially overlapping window across the sequence. This dual recursion allows us to better handle viewpoint shifts and extended sequences while preserving consistent geometry and appearance throughout the generated views.

\subsubsection{Spatial Auto-Regressive Expansion} 
\refine{A direct warp to a distant target pose suffers from depth noise and heavy occlusions, providing weak cues for inpainting. We address this by decomposing the overall viewpoint shift into a chain of small, well-conditioned hops. Following ViewCrafter \cite{yu2024viewcrafter}, we spline-interpolate a sequence of $P$ virtual cameras that smoothly bridge the source and target views. Starting at the source pose, we iterate through this chain: each step we warp the current video with estimated depth to the next virtual pose, inpaint the resulting holes, then re-estimate depth and align its scale to the previous depth using the global solver of \cite{wang2024shape,ma2025see3d}. The refined clip produced at step $p$ becomes the input for step $p{+}1$. Because every hop spans only a few degrees, projection error and missing regions remain modest, allowing the diffusion model to repair artifacts thoroughly before the next warp. After $P$ iterations, the process converges at the true target view, yielding high-quality frames for viewpoint shifts while preserving appearance and geometry throughout the traversal.}

\subsubsection{Temporal Auto-Regressive Inference}
\refine{Hardware limits restrict each diffusion pass to a short clip, yet the final output must remain coherent over a much longer horizon. We address this with a sliding-window scheme.
Let $L$ be the window length and $m<L$ the overlap. We first denoise the initial \(L\) frames. For each subsequent window, we append the last \(m\) clean latent frames from the previous output to the next \(L-m\) noisy frames, forming an \(L\)-frame batch for denoising. The shared \(m\)-frame segment acts as a temporal anchor, enforcing both appearance and motion continuity and eliminating seam artifacts. Iterating the process yields long, temporally coherent videos without increasing memory usage. Spatial expansion decomposes viewpoint shifts into well-conditioned steps, while temporal recursion amortizes memory and propagates dynamics across windows. Together, they deliver temporally consistent, pose-free novel views at viewpoint shifts and long sequences, which are capabilities unattainable by prior pipelines that operate in a single pass.}
\section{Experiments}
\label{sec:experiments}

\begin{table}[t]
\centering
\caption{\textbf{Comparative study with SoTA methods for 4D scene reconstruction} on the five different scenes from the iPhone \cite{iphone} dataset. Symbols ``$\uparrow$''~/~``$\downarrow$'' denote scores that are the ``higher''~/~``lower'' the better.}
\resizebox{\linewidth}{!}{
\begin{tabular}{r|r|ccccc|c}
    \toprule
    \textbf{Method} & \textbf{Venue} & Apple & Block & Paper & Spin & Teddy & \textbf{Avg}
    \\\midrule\midrule
    \rowcolor{see_purple!15}\multicolumn{8}{c}{\textbf{Metric: PSNR} $\uparrow$}
    \\
    GCD \cite{hoorick2024gcd} & \textcolor{gray}{ECCV'24} & $9.82$ & $12.30$ & $9.75$ & $10.37$ & $11.61$ & $10.77$
    \\
    ViewCrafter \cite{yu2024viewcrafter} & \textcolor{gray}{TPAMI'25} & $10.19$ & $10.28$ & $10.63$ & $11.15$ & $11.50$ & $10.75$
    \\
    Shape-of-Motion \cite{wang2024shape} & \textcolor{gray}{ICCV '25} & $11.06$ & $11.72$ & $11.93$ & $11.28$ & $10.42$ & $11.28$
    \\
    DaS \cite{gu2025diffusion} & \textcolor{gray}{Siggraph'25} & $10.02$ & $11.64$ & $10.27$ & $11.11$ & $11.82$ & $10.97$
    \\
    ReCamMaster \cite{bai2025recammaster} & \textcolor{gray}{ICCV'25} & $10.96$ & $12.67$ & $11.88$ & $12.25$ & $12.37$ & $12.02$
    \\
    TrajectoryCrafter \cite{yu2025trajectorycrafter} & \textcolor{gray}{ICCV'25} & $13.88$ & $14.21$ & $14.89$ & $14.51$ & $13.73$ & $14.24$
    \\
    \textbf{\ours} & \textbf{Ours} & $\mathbf{13.98}$ & $\mathbf{14.67}$ & $\mathbf{15.24}$ & $\mathbf{14.72}$ & $\mathbf{14.22}$ & $\mathbf{14.56}$
    \\\midrule
    \rowcolor{see_green!15}\multicolumn{8}{c}{\textbf{Metric: SSIM} $\uparrow$}
    \\
    GCD \cite{hoorick2024gcd} & \textcolor{gray}{ECCV'24} & $0.215$ & $0.458$ & $0.398$ & $0.324$ & $0.385$ & $0.356$
    \\
    ViewCrafter \cite{yu2024viewcrafter} & \textcolor{gray}{TPAMI'25} & $0.245$ & $0.427$ & $0.344$ & $0.308$ & $0.372$ & $0.339$
    \\
    Shape-of-Motion \cite{wang2024shape} & \textcolor{gray}{ICCV'25} & $0.197$ & $0.446$ & $0.425$ & $0.319$ & $0.357$ & $0.349$
    \\
    DaS \cite{gu2025diffusion} & \textcolor{gray}{Siggraph'25} & $0.217$ & $0.388$ & $0.356$ & $0.312$ & $0.381$ & $0.331$
    \\
    ReCamMaster \cite{bai2025recammaster} & \textcolor{gray}{ICCV'25} & $0.264$ & $0.454$ & $0.471$ & $0.344$ & $0.401$ & $0.387$
    \\
    TrajectoryCrafter \cite{yu2025trajectorycrafter} & \textcolor{gray}{ICCV'25} & $0.285$ & $0.528$ & $0.482$ & $0.380$ & $0.411$ & $0.417$
    \\
    \textbf{\ours} & \textbf{Ours} & $\mathbf{0.309}$ & $\mathbf{0.555}$ & $\mathbf{0.514}$ & $\mathbf{0.399}$ & $\mathbf{0.434}$ & $\mathbf{0.442}$
    \\\midrule
    \rowcolor{see_yellow!15}\multicolumn{8}{c}{\textbf{Metric: LPIPS} $\downarrow$}
    \\
    GCD \cite{hoorick2024gcd} & \textcolor{gray}{ECCV'24} & $0.738$ & $0.590$ & $0.535$ & $0.576$ & $0.629$ & $0.614$
    \\
    ViewCrafter \cite{yu2024viewcrafter} & \textcolor{gray}{TPAMI'25} & $0.750$ & $0.615$ & $0.521$ & $0.533$ & $0.606$ & $0.605$
    \\
    Shape-of-Motion \cite{wang2024shape} & \textcolor{gray}{ICCV'25} & $0.879$ & $0.601$ & $0.486$ & $0.560$ & $0.650$ & $0.635$
    \\
    DaS \cite{gu2025diffusion} & \textcolor{gray}{Siggraph'25} & $0.732$ & $0.593$ & $0.520$ & $0.551$ & $0.608$ & $0.601$
    \\
    ReCamMaster \cite{bai2025recammaster} & \textcolor{gray}{ICCV'25} & $0.683$ & $0.537$ & $0.491$ & $0.545$ & $0.572$ & $0.566$
    \\
    TrajectoryCrafter \cite{yu2025trajectorycrafter} & \textcolor{gray}{ICCV'25} & $0.612$ & $0.479$ & $0.471$ & $0.518$ & $0.513$ & $0.519$
    \\
    \textbf{\ours} & \textbf{Ours} & $\mathbf{0.581}$ & $\mathbf{0.455}$ & $\mathbf{0.439}$ & $\mathbf{0.501}$ & $\mathbf{0.486}$ & $\mathbf{0.492}$
    \\
    \bottomrule
\end{tabular}}
\label{tab:class_iphone}
\end{table}
\begin{table}[t]
\centering
\caption{\textbf{Comparative study with SoTA methods for video generation} following VBench \cite{vbench} evaluation protocol. The metrics measure video and frame qualities and consistencies. All scores are the higher the better.}
\resizebox{\linewidth}{!}{
\begin{tabular}{r|ccccccc}
    \toprule
    \multirow{2}{*}{\textbf{Method}} & \cellcolor{see_purple!15}\textbf{Subj.} & \cellcolor{see_purple!15}\textbf{Back.} & \cellcolor{see_green!15}\textbf{Temp.} & \cellcolor{see_green!15}\textbf{Motion} & \cellcolor{see_yellow!15}\textbf{Image.} & \cellcolor{see_yellow!15}\textbf{Aesth.} 
    \\
    & \cellcolor{see_purple!15}\textbf{Consist} & \cellcolor{see_purple!15}\textbf{Consist} & \cellcolor{see_green!15}\textbf{Flick} & \cellcolor{see_green!15}\textbf{Smooth} & \cellcolor{see_yellow!15}\textbf{Quality} & \cellcolor{see_yellow!15}\textbf{Quality} 
    \\\midrule\midrule
    DaS \cite{gu2025diffusion} & $89.44$ & $91.69$ & $96.11$ & $95.58$ & $50.64$ & $37.49$ 
    \\
    ReCamMaster \cite{bai2025recammaster} & $90.56$ & $93.42$ & $95.11$ & $\mathbf{98.12}$ & $52.47$ & $39.65$
    \\
    TrajectoryCrafter \cite{yu2025trajectorycrafter} & $89.61$ & $92.55$ & $92.78$ & $93.49$ & $52.20$ & $37.62$
    \\
    \textbf{\ours} & $\mathbf{92.18}$ & $\mathbf{94.63}$ & $\mathbf{96.66}$ & $97.87$ & $\mathbf{53.15}$ & $\mathbf{41.35}$
    \\
    \bottomrule
\end{tabular}}
\label{tab:vbench}
\end{table}
\begin{figure}[t]
    \centering
    \includegraphics[width=\linewidth]{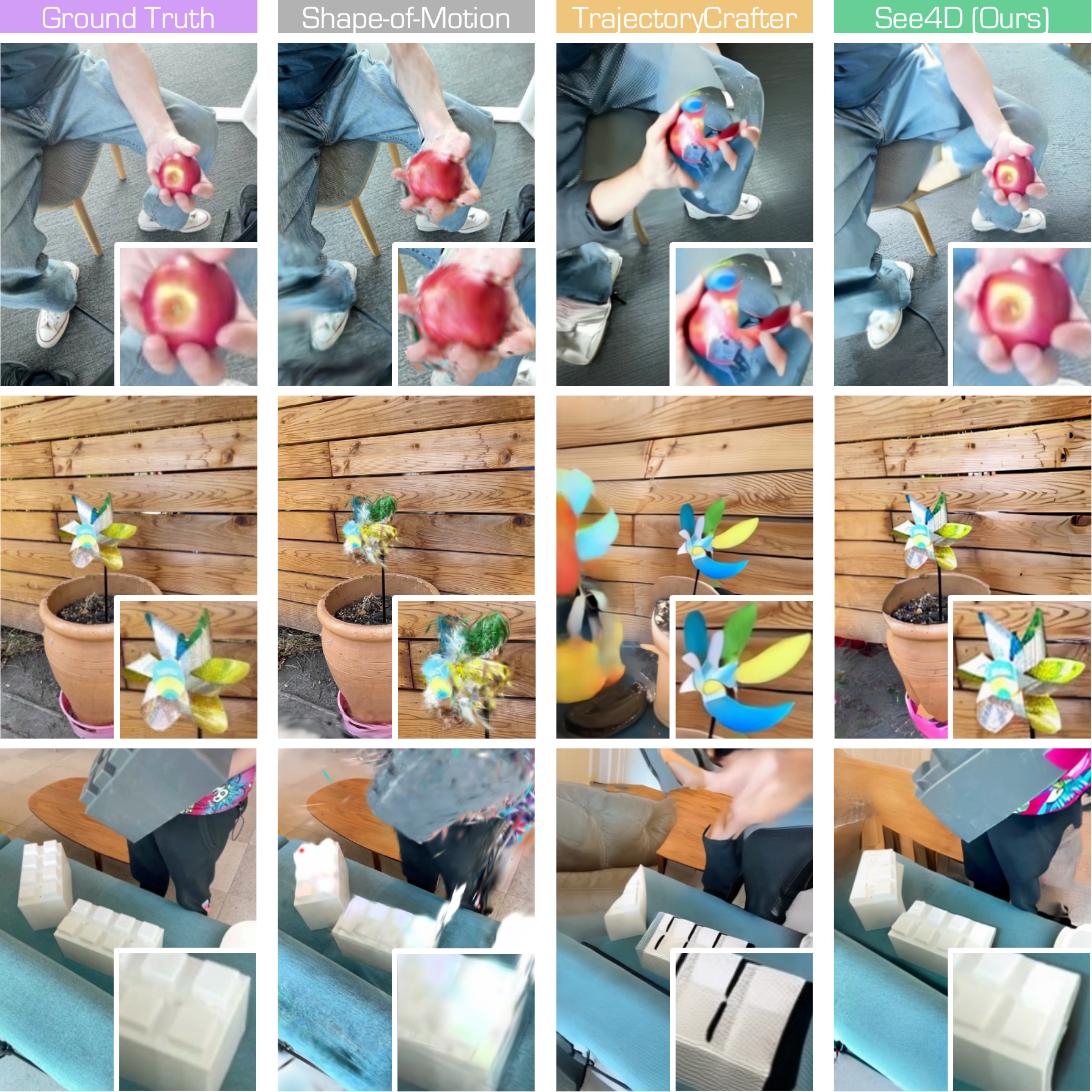}
    \caption{\textbf{Qualitative comparisons} of \ours~and baselines for 4D reconstruction on the iPhone \cite{iphone} dataset.}
    \vspace{-0.2cm}
    \label{fig:qual_recon}
\end{figure}
\begin{figure*}[t]
    \centering
    \includegraphics[width=\linewidth]{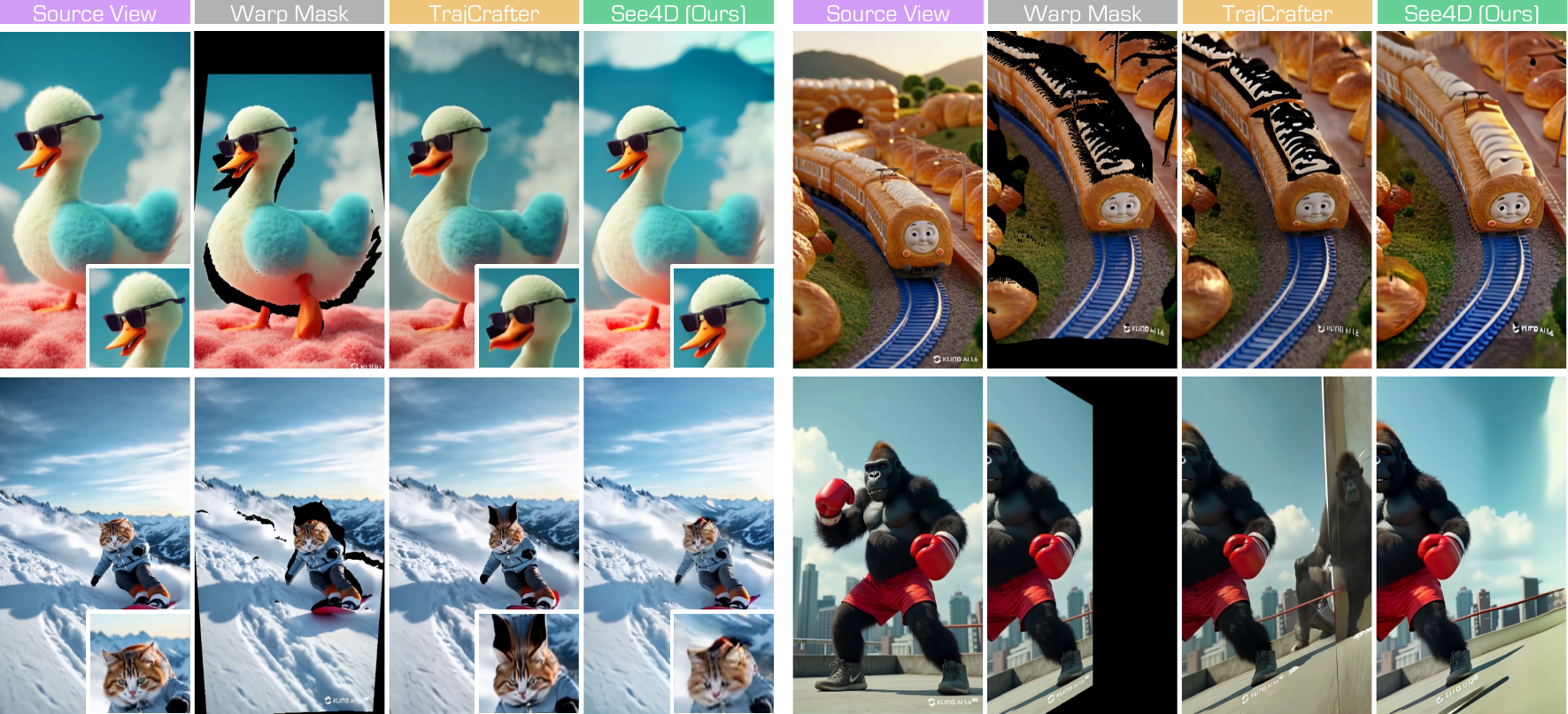}
    \caption{\textbf{Qualitative comparisons} of TrajectoryCrafter~\cite{ji2025posetraj} and \ours~for 4D generation. A single frame from each example is shown due to space limits. 
    Kindly refer to our \textbf{Appendix File} for high-resolution demo videos and frame-by-frame examples.}
    \label{fig:qual_gen}
\end{figure*}

\subsection{Experimental Settings}
\subsubsection{Datasets}
We train on two synthetic multi-view dynamic datasets with synchronized captures for 4D consistency. \textit{SynCamMaster}~\cite{bai2024syncammaster} provides $34$K clips from $3.4$K scenes filmed by ten static cameras. \textit{ReCamMaster}~\cite{bai2025recammaster} adds motion, with $13.6$K scenes captured by nine moving and one fixed camera, yielding $136$K clips. Altogether, $170$K temporally aligned videos cover diverse baselines, dynamics, environments, and trajectories. During training, we sample two clips (source/target) per SynCamMaster scene and the moving plus static clip per ReCamMaster scene.

\subsubsection{Implementation Details}
We initialize from See3D weights~\cite{ma2025see3d}. Training uses \(512 \times 512\) resolution, $16$-frame sequences, batch size $16$, and $10$K iterations at  \(1 \times 10^{-5}\) learning rate. We employ bfloat16 precision and classifier-free guidance~\cite{ho2022classifier} by dropping visual conditions with $0.1$ probability. Training runs on eight NVIDIA A800-80GB GPUs for approximately $48$ hours. At inference, we use a DDIM sampler~\cite{song2020denoising} with the same guidance strategy.

\subsubsection{Evaluation Protocol}

We evaluate our method with two primary settings: 4D reconstruction and cross-view video generation.

\noindent{\textbf{4D Reconstruction.}}
We assess monocular 4D reconstruction on the public iPhone dataset~\cite{iphone}. Following~\cite{yu2025trajectorycrafter}, we select five handheld sequences, where the moving video serves as the source input, and the first synchronized fixed-camera clip as the target. Each scene is furnished with COLMAP-derived poses and aligned depth maps from~\cite{yu2025trajectorycrafter, wang2024shape}, enabling rigorous geometric evaluation. Reconstruction quality is measured by PSNR, SSIM, and LPIPS~\cite{zhang2018unreasonable}.

\noindent{\textbf{Cross-View Video Generation.}}
To more thoroughly evaluate our model’s ability to synthesize a fixed target view from a single handheld clip, we assemble an in-the-wild dataset of 200 monocular videos drawn from WebVid~\cite{bain2021frozen}. For each clip, we randomly select a static camera viewpoint as the target, spanning pan, tilt, and hemispherical arc trajectories around the actor’s \(x\)- and \(y\)-axes. Quality is assessed using the VBench protocol~\cite{vbench}.

\noindent{\textbf{Comparison Methods.}}  
We compare against serval state-of-the-art methods: GCD~\cite{hoorick2024gcd}, ViewCrafter~\cite{yu2024viewcrafter}, DaS~\cite{gu2025diffusion}, TrajectoryCrafter~\cite{yu2025trajectorycrafter}, ReCamMaster~\cite{bai2025recammaster}, and Shape-of-Motion~\cite{wang2024shape}. The first five are diffusion-based generative models, while Shape-of-Motion is the leading reconstruction-oriented 4D approach. Since ReCamMaster and DaS do not target reconstruction objectives, their reconstruction metrics are reported only for qualitative reference.

\subsection{Comparative Study}
\subsubsection{4D Reconstruction}
Tab. \ref{tab:class_iphone} shows that \ours~ attains the best average PSNR, SSIM, and LPIPS on all five iPhone scenes, with consistent gains per sequence, showing that our \textit{trajectory-to-camera} formulation handles both rigid and deformable content. ReCamMaster relies on explicit pose input and fails outside its training frame, whereas our depth-warp conditioning needs no poses. ViewCrafter, designed for static images, loses 4D coherence when objects move, while our spatial–temporal transformer preserves alignment across time. TrajectoryCrafter and DaS assume clean geometric cues and degrade under noise, a weakness mitigated by our noise-adaptive condition. Shape-of-Motion leaves large voids when points disappear, but our warp synthesis and inpainting fill such occlusions. 

\subsubsection{Cross-View Video Generation}  
We evaluate on VBench’s generation protocol \cite{vbench} in Tab.~\ref{tab:vbench}, where baselines leverage large, dedicated video-diffusion backbones.  Despite our core network being a multi-view synthesis model, \ours~ leads on five of six metrics and ranks a close second on motion smoothness.  These results underscore the power of our view-conditional inpainting model and spatial–temporal auto-regressive inference pipeline. By fusing explicit geometry cues with robust temporal priors, our method exceeds the performance of models tailored solely for video generation.

\begin{table}[t]
\centering
\caption{\textbf{Ablation studies of proposed components} on iPhone \cite{iphone}. Symbols ``$\uparrow$''~/~``$\downarrow$'' denote scores that are the ``higher''~/~``lower'' the better.}
\resizebox{\linewidth}{!}{
\begin{tabular}{c|l|ccc}
    \toprule
    {\textbf{Module}} & \textbf{Component} & \cellcolor{see_purple!15}\textbf{PSNR} $\uparrow$ & \cellcolor{see_green!15}\textbf{SSIM} $\uparrow$ & \cellcolor{see_yellow!15}\textbf{LPIPS} $\downarrow$
    \\\midrule\midrule
    \multirow{4}{*}{\begin{tabular}[c]{@{}c@{}}\textbf{Warp Syn.}\\ {(\textit{cf.} Sec.\ref{sec:Realistic Warp Synthesis})}\end{tabular}} & from See3D \cite{ma2025see3d} & $11.93$ & $0.377$ & $0.568$
    \\
    & from Vivid4D \cite{huang2025vivid4d}
    & $12.23$ & $0.391$ & $0.556$
    \\
    & from TrajectoryCrafter \cite{yu2025trajectorycrafter} & $12.48$ & $0.420$ & $0.537$
    \\
    & \textbf{Ours} & $14.56$ & $0.442$ & $0.492$
    \\\midrule
    \multirow{2}{*}{\begin{tabular}[c]{@{}c@{}}\textbf{Noise Ada.}\\ {(\textit{cf.} Sec.\ref{sec:Noise‑Adaptive Condition})}\end{tabular}} & w/o noise adaptive condition & $13.47$ & $0.411$ & $0.508$
    \\
    & \textbf{Ours} & $14.56$ & $0.442$ & $0.492$
    \\\midrule
    \multirow{2}{*}{\begin{tabular}[c]{@{}c@{}}\textbf{Spa.-Tem.}\\ {(\textit{cf.} Sec.\ref{sec:Spatial–Temporal Backbone})}\end{tabular}} & w/o spatial-temporal transformer & $10.66$ & $0.347$ & $0.685$
    \\
    & \textbf{Ours} & $14.56$ & $0.442$ & $0.492$
    \\\midrule
    \multirow{3}{*}{\begin{tabular}[c]{@{}c@{}}\textbf{Auto.-Reg.}\\ {(\textit{cf.} Sec.\ref{sec:inference})}\end{tabular}} & w/o spatial regression  & $13.58$ & $0.424$ & $0.518$
    \\
    & w/o temporal regression  & $13.02$ & $0.438$ & $0.517$
    \\
    & \textbf{Ours} & $14.56$ & $0.442$ & $0.492$
    \\
    \bottomrule
\end{tabular}}
\label{tab:ablation}
\end{table}
\begin{figure*}[t]
    \centering
    \includegraphics[width=\linewidth]{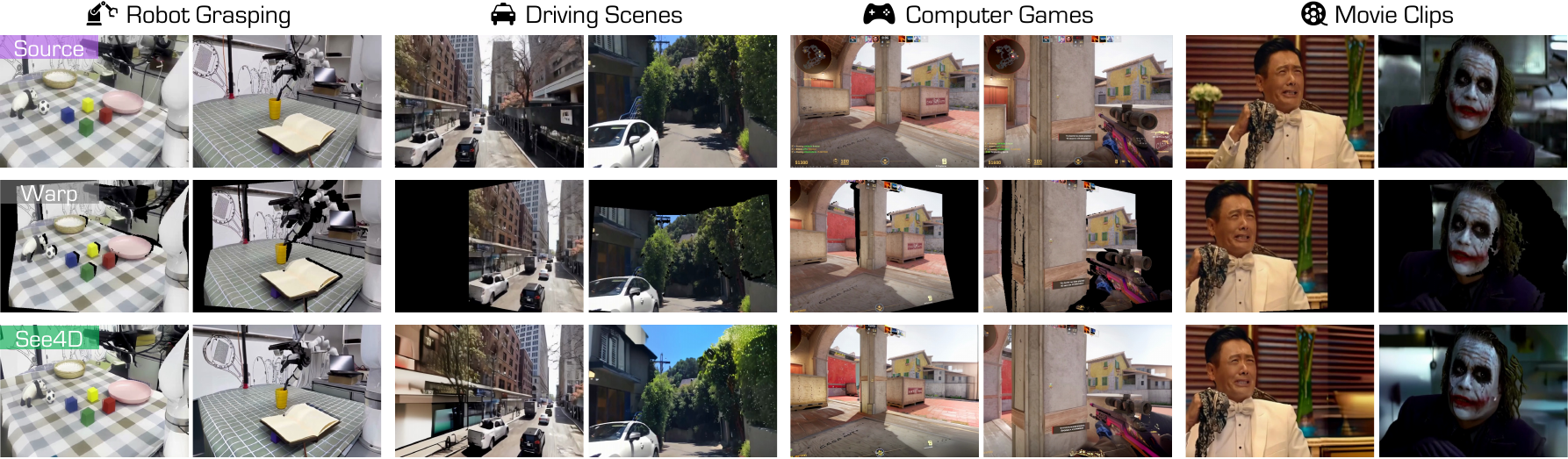}
    \caption{\textbf{Illustrative examples} of \ours~for various applications (robotics, driving, games, movies, \etc). A single frame from each example is shown due to space limits. 
    Kindly refer to our \textbf{Appendix File} for high-resolution demo videos and frame-by-frame examples.}
    \label{fig:tasks}
\end{figure*}

\subsubsection{Qualitative Assessment} 
\begin{figure*}[t]
\vspace{0.2cm}
    \centering
    \includegraphics[width=\linewidth]{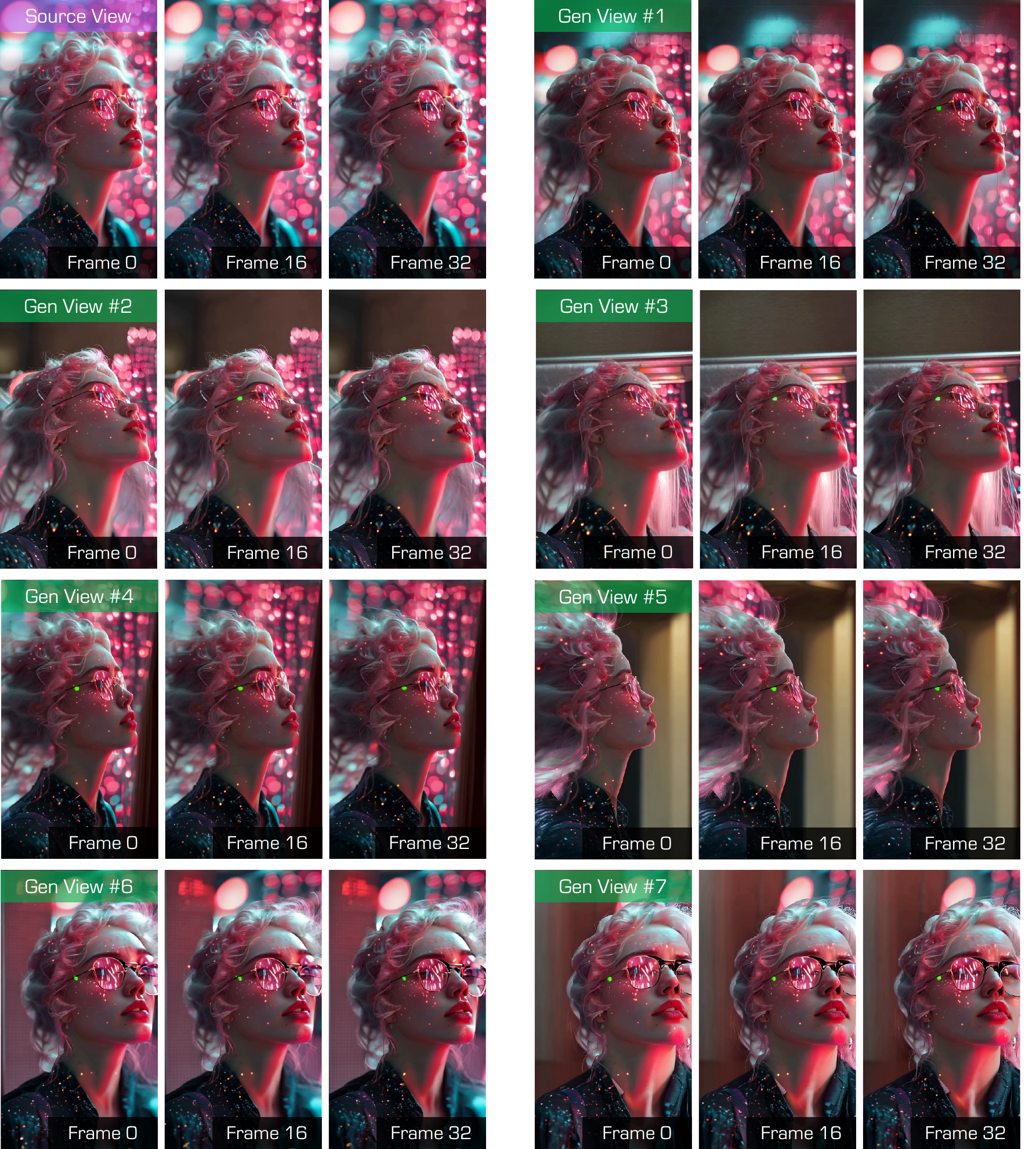}
    \caption{Generated examples of \ours~on 4D generation. The source video is from KLing. Kindly refer to our
    \textbf{Appendix File} for high-resolution demo videos and frame-by-frame examples.}
    \label{fig:two_pages_1}
\end{figure*}

Qualitative comparisons in Fig.~\ref{fig:qual_recon} on iPhone reconstruction show that our method recovers sharp geometry and stable parallax with less flicker, outperforming Shape-of-Motion’s blurring and TrajectoryCrafter’s minor inconsistencies. Fig.~\ref{fig:qual_gen} on cross-view generation demonstrates smooth, occlusion-aware texture synthesis and seamless transitions, while TrajectoryCrafter exhibits bleed-through and jitter. Additional cross-view generation examples are provided in Fig.~\ref{fig:two_pages_1}. Please refer to our supplementary materials for more results and video demonstrations.

\subsection{Ablation Study}
Tab.~\ref{tab:ablation} confirms that each component of our design is essential.  Substituting our \textit{realistic warp synthesis} with the simpler schemes of prior work causes a substantial drop in all reconstruction metrics, demonstrating that modeling real-world occlusions and depth noise is critical to provide a reliable view prior. Omitting \textit{noise-adaptive condition} degrades performance across the board, showing that dynamically scaling perturbation to warp quality is key to preventing over-reliance on imperfect geometric cues and to encouraging robust inpainting. Removing the \textit{spatial–temporal backbone} reduces reconstruction fidelity dramatically, underscoring the necessity of joint cross-frame attention for preserving coherent appearance in dynamic video content. Ablating either the spatial expansion or the temporal recursion in our \textit{auto-regressive inference} pipeline leads to noticeable quality declines, illustrating that both progressive view hops and overlapping diffusion windows are indispensable for high-quality novel-view synthesis.

\subsection{Downstream Applications}
\begin{figure*}[t]
\vspace{0.2cm}
    \centering
    \includegraphics[width=\linewidth]{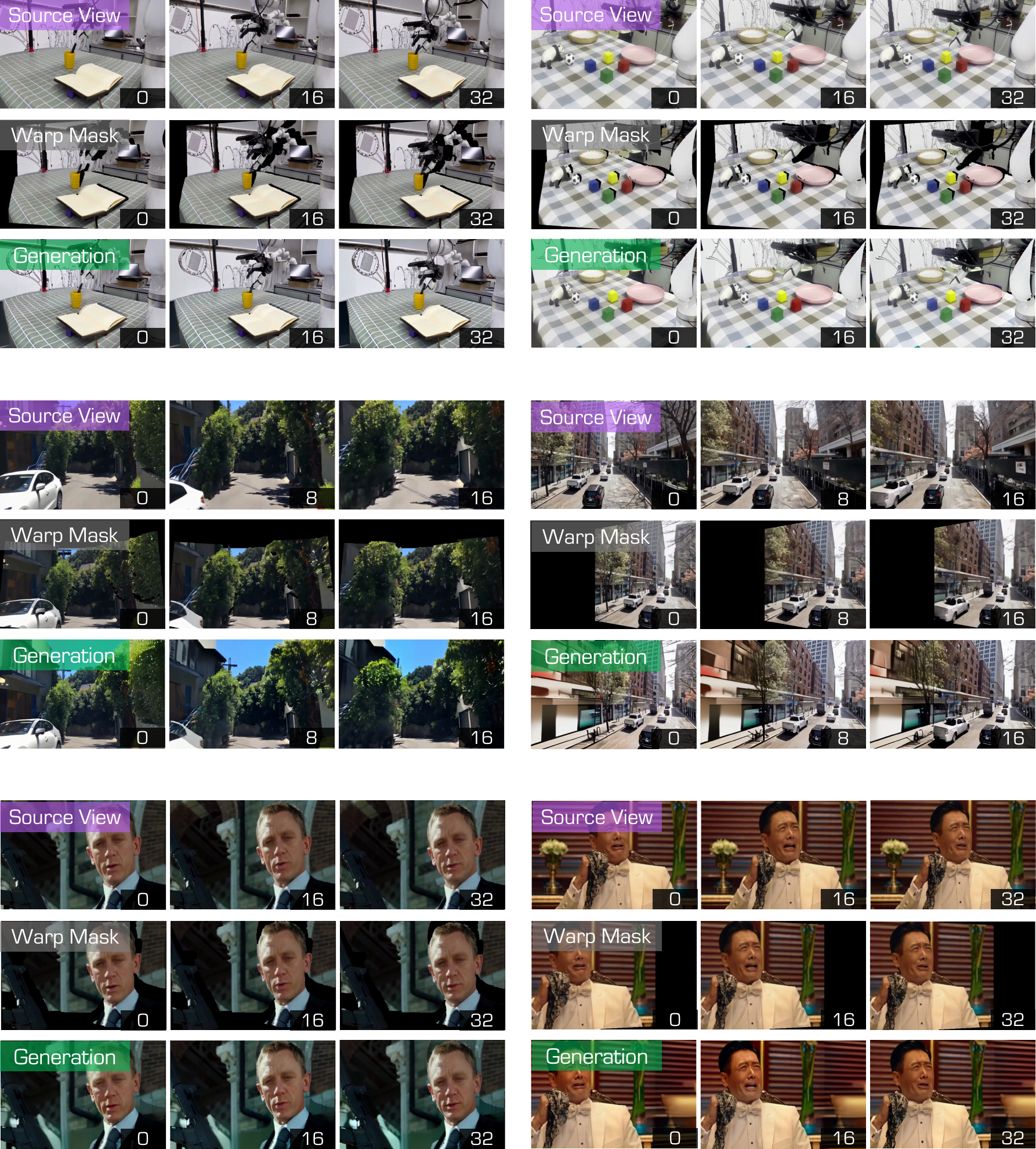}
    \caption{Generated examples of \ours~on 4D generation. The source videos are from Video Prediction Policy, Vista, and ReCamMaster. 
    Kindly refer to our \textbf{Appendix File} for high-resolution demo videos and frame-by-frame examples.}
    \label{fig:two_pages_2}
\end{figure*}

Beyond 4D reconstruction, \ours~enables synchronized multi-view video synthesis across diverse applications, as illustrated in Fig.~\ref{fig:tasks} and Fig.~\ref{fig:two_pages_2}. In \textit{robotic manipulation}, it generates a panoramic overview of the workbench for grasp planning. In \textit{autonomous driving}, it augments dash-cam footage with virtual side- and rear-view streams that faithfully reproduce static infrastructure and dynamic traffic participants. In \textit{interactive gaming}, single in-game captures yield multi-angle “fly-through” replays preserving character and environment dynamics. In \textit{cinematic post-production}, it produces stabilized off-axis takes from handheld clips, filling occlusions, and maintaining fidelity for seamless re-framing. 
\section{Conclusion}
\label{sec:conclusion}

In this work, we introduced \ours, a pose-free framework that converts a single hand-held video into a bank of synchronized, fixed-view sequences through a trajectory-to-camera formulation. Conditioning a latent diffusion model on depth-warped images and masks removes the need for explicit 6-DoF annotations while preserving geometric cues. Unlike trajectory-to-trajectory methods, our formulation decouples scene modeling from camera motion, yielding more stable synthesis. A view-conditional inpainting model enforces cross-view and cross-frame consistency, while a spatial–temporal auto-regressive pipeline extends generation across viewpoints and long sequences, producing coherent videos that can be fused into dense 4D modeling and representation. Experiments show improved cross-view quality and reconstruction accuracy over existing baselines, and we demonstrate benefits across applications such as robotic manipulation, autonomous driving, interactive gaming, and cinematic post-production.

\clearpage

\clearpage
\clearpage\clearpage

\begin{figure*}[t]
\vspace{0.16cm}
    \centering
    \includegraphics[width=\linewidth]{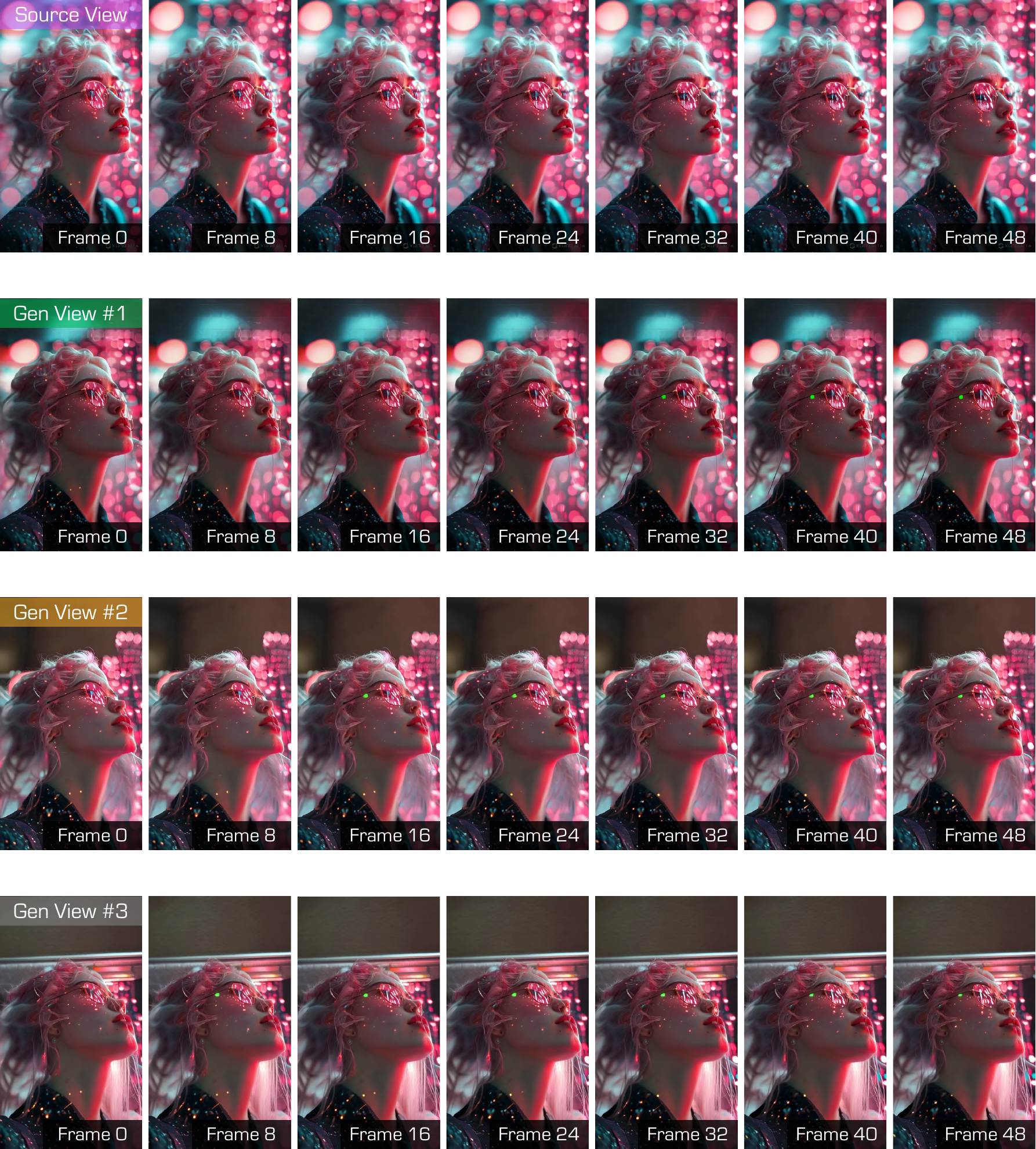}
    \caption{Generated examples (\#1, \#2, \#3) of \ours~on 4D generation. The source video is from KLing.}
    \label{fig:teaser_full_1}
\end{figure*}

\clearpage\clearpage
\begin{figure*}[t]
\vspace{0.16cm}
    \centering
    \includegraphics[width=\linewidth]{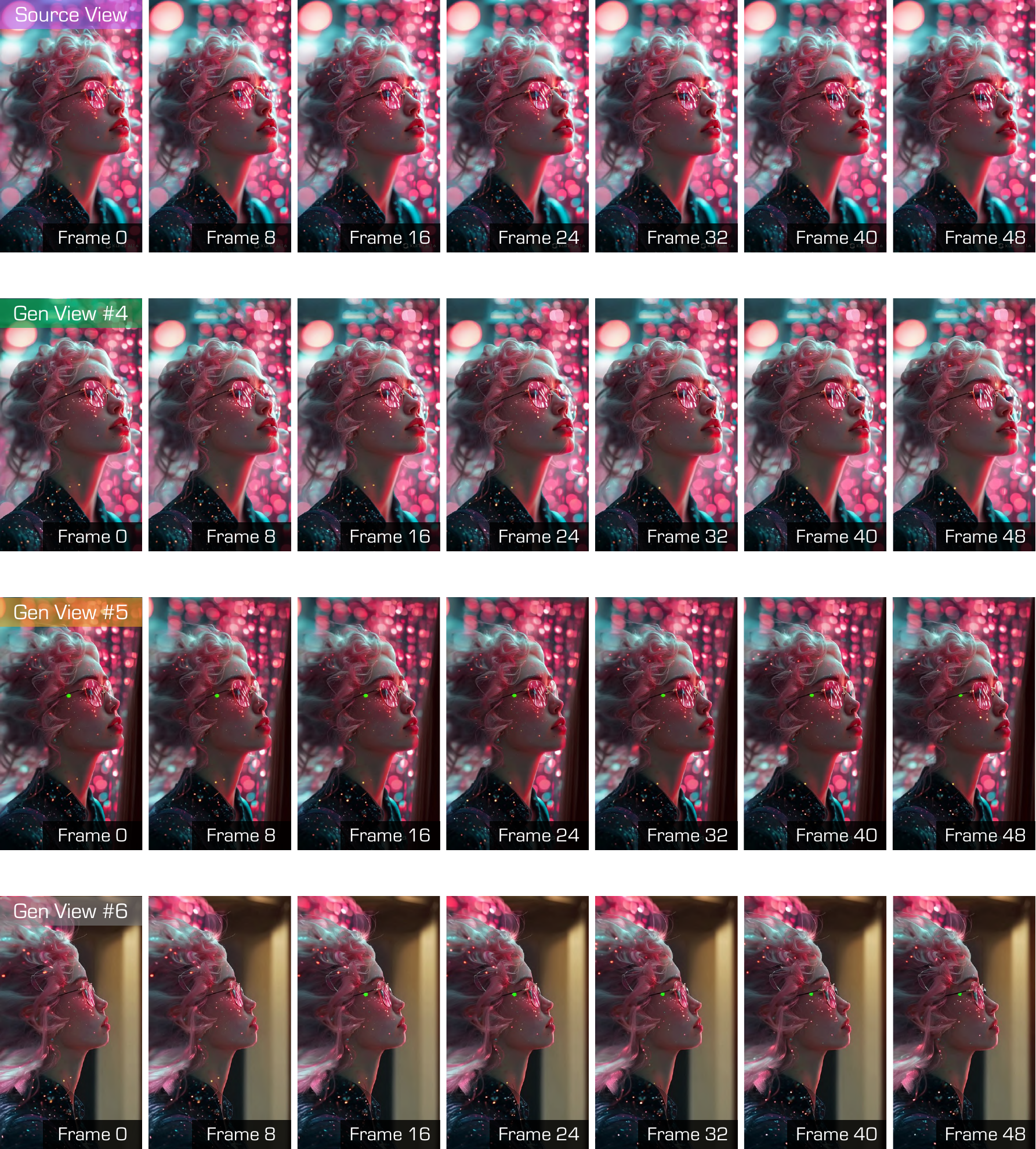}
    \caption{Generated examples (\#4, \#5, \#6) of \ours~on 4D generation. The source video is from KLing.}
    \label{fig:teaser_full_2}
\end{figure*}

\clearpage\clearpage
\begin{figure*}[t]
\vspace{0.16cm}
    \centering
    \includegraphics[width=\linewidth]{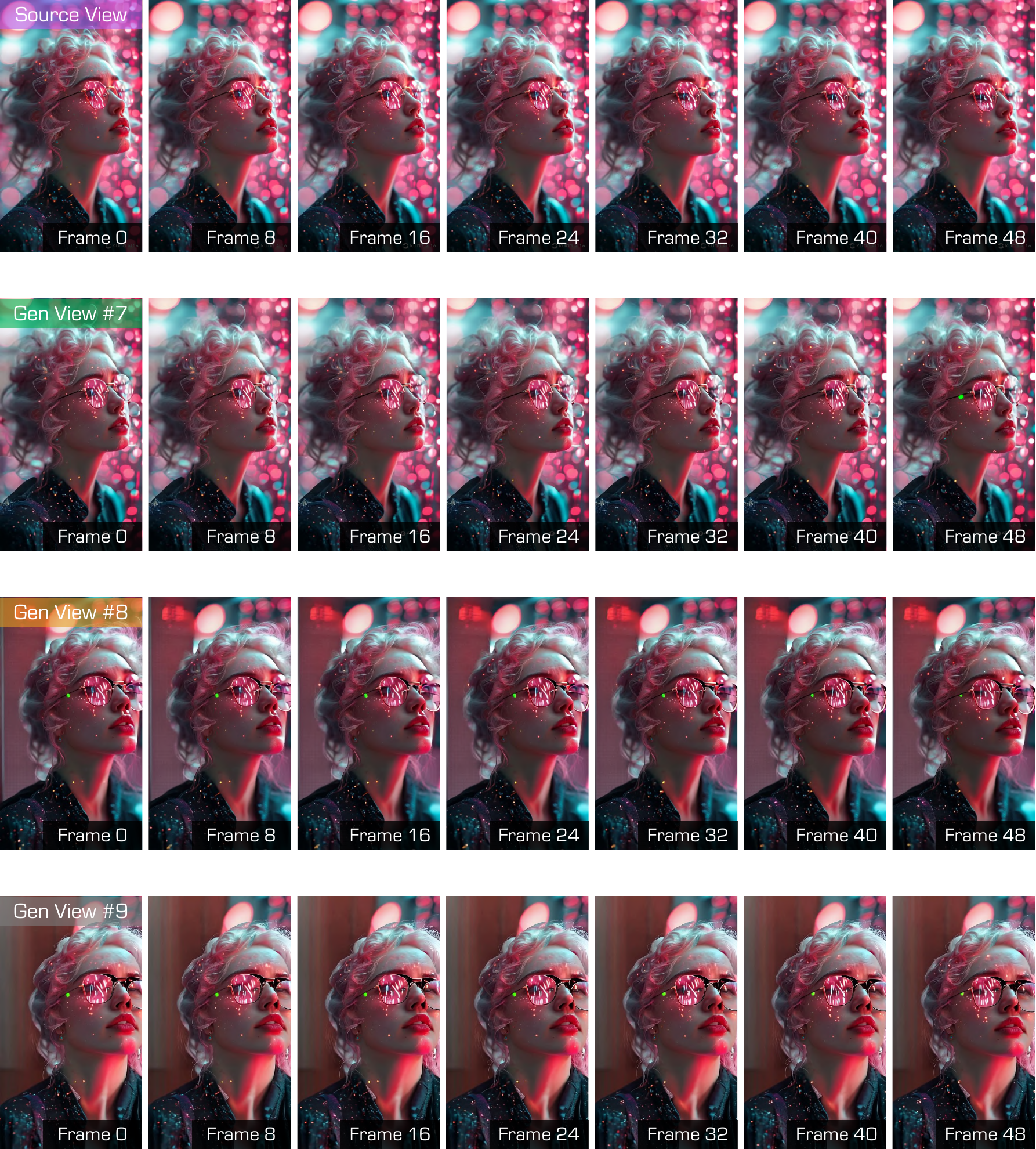}
    \caption{Generated examples (\#7, \#8, \#9) of \ours~on 4D generation. The source video is from KLing.}
    \label{fig:teaser_full_3}
\end{figure*}

\clearpage\clearpage
\begin{figure*}[t]
\vspace{0.16cm}
    \centering
    \includegraphics[width=\linewidth]{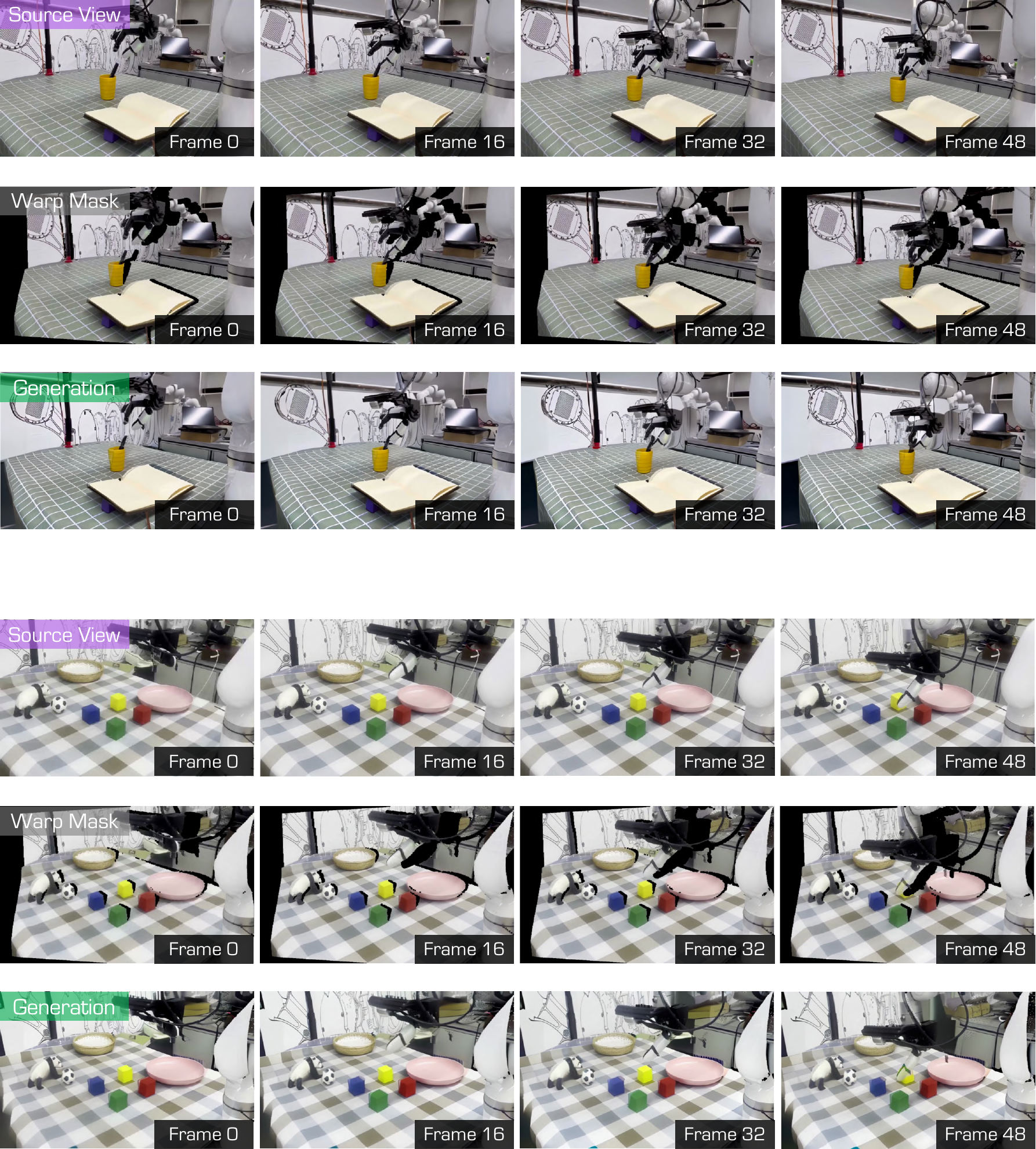}
    \caption{Application use cases of \ours~on robot grasping. The source videos are from Video Prediction Policy.}
    \label{fig:qua_robotics_1}
\end{figure*}

\clearpage\clearpage
\begin{figure*}[t]
\vspace{0.16cm}
    \centering
    \includegraphics[width=\linewidth]{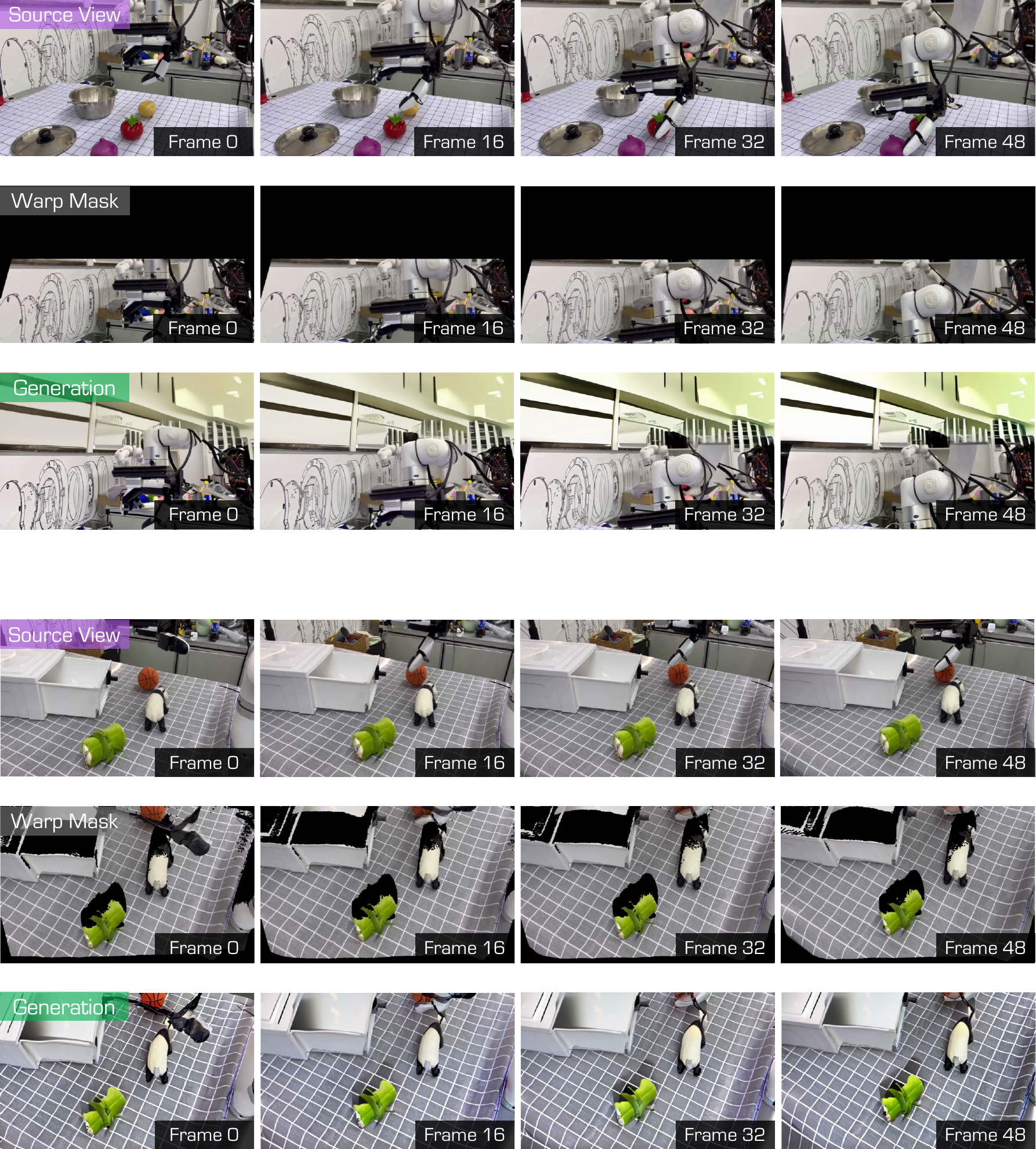}
    \caption{Application use cases of \ours~on robot grasping. The source videos are from Video Prediction Policy.}
    \label{fig:qua_robotics_2}
\end{figure*}

\clearpage\clearpage
\begin{figure*}[t]
\vspace{0.16cm}
    \centering
    \includegraphics[width=\linewidth]{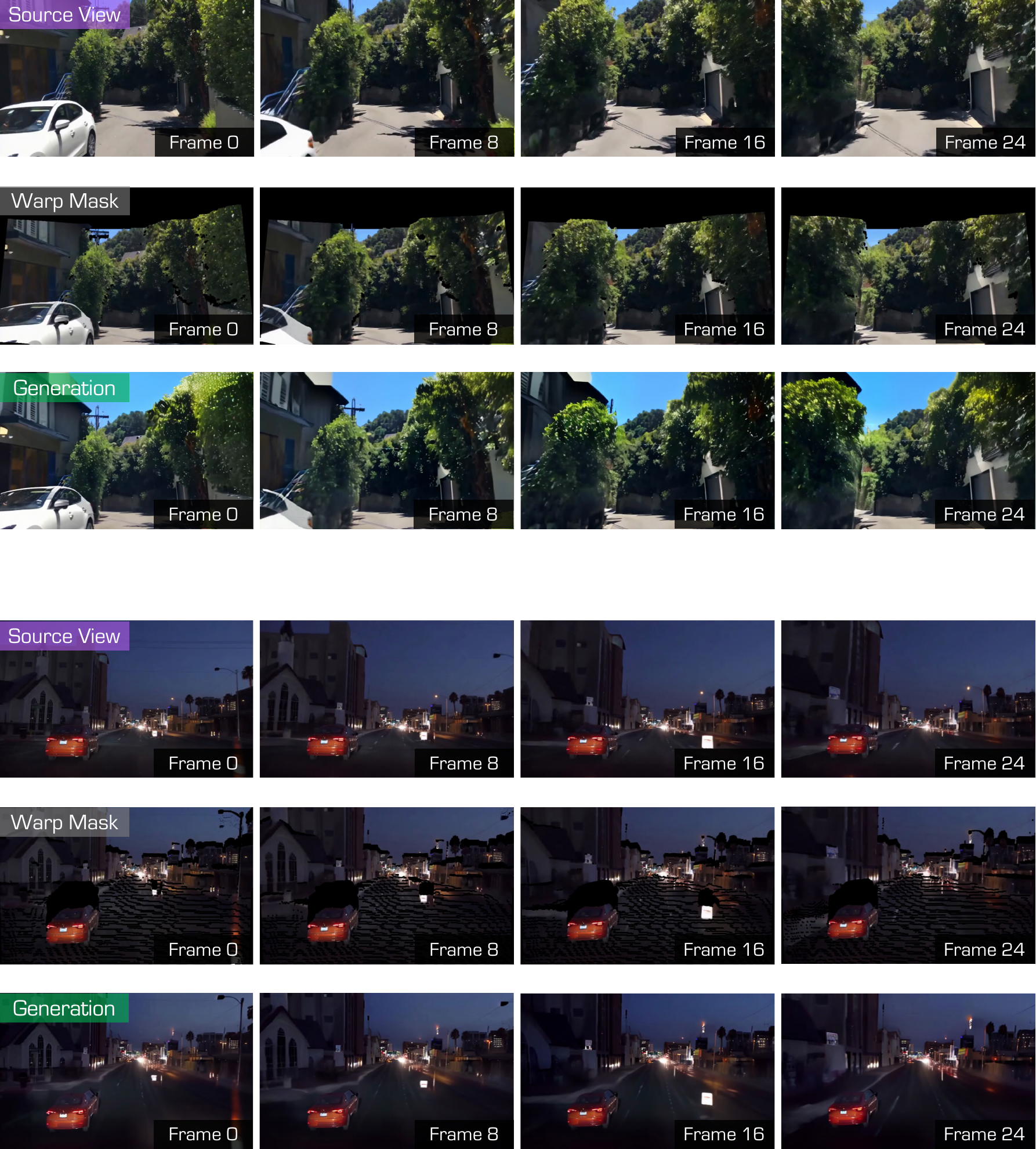}
    \caption{Application use cases of \ours~on driving scene generation. The source videos are from Vista.}
    \label{fig:qua_driving_1}
\end{figure*}

\clearpage\clearpage
\begin{figure*}[t]
\vspace{0.16cm}
    \centering
    \includegraphics[width=\linewidth]{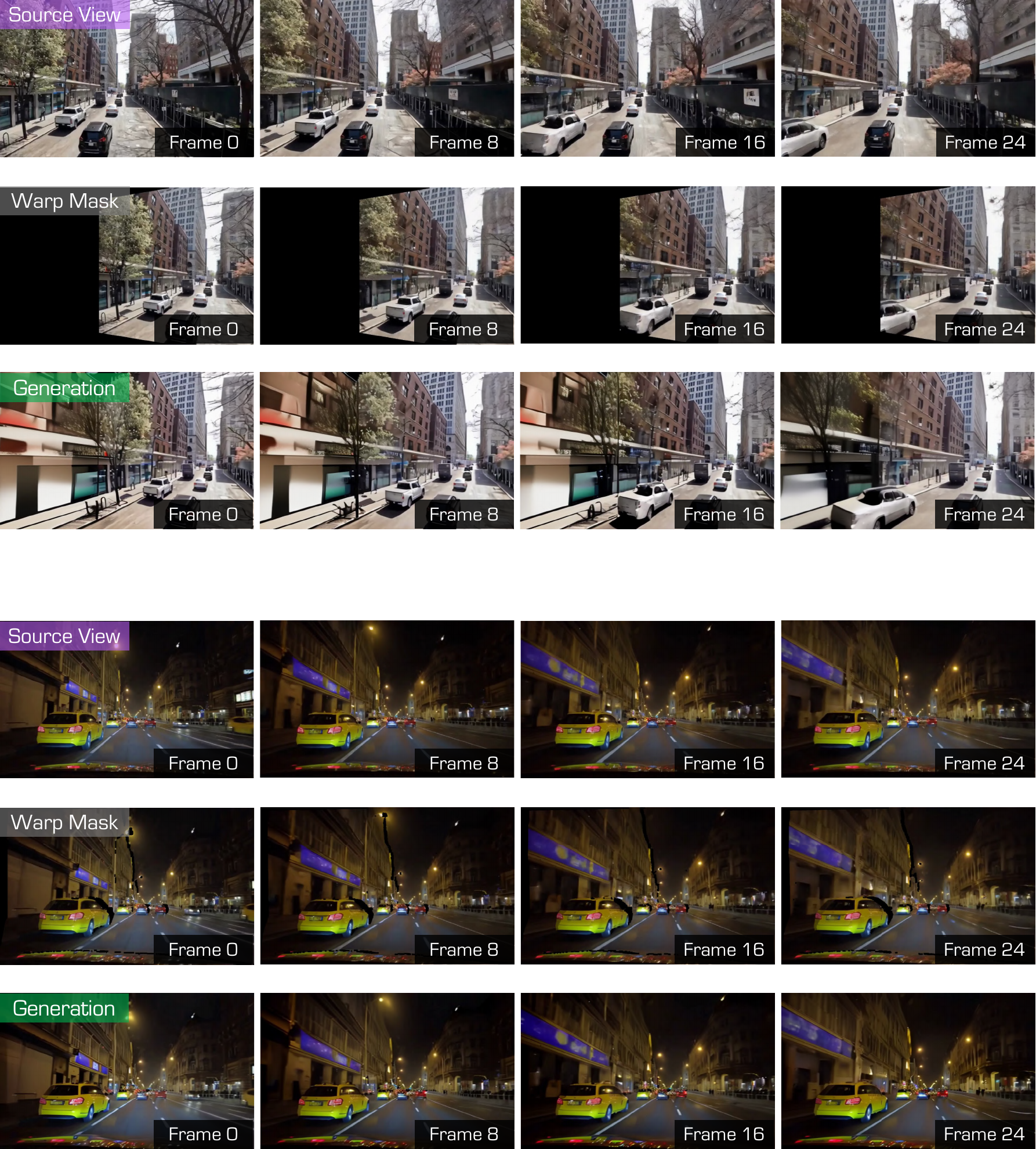}
    \caption{Application use cases of \ours~on driving scene generation. The source videos are from Vista.}
    \label{fig:qua_driving_2}
\end{figure*}

\clearpage\clearpage
\begin{figure*}[t]
\vspace{0.16cm}
    \centering
    \includegraphics[width=\linewidth]{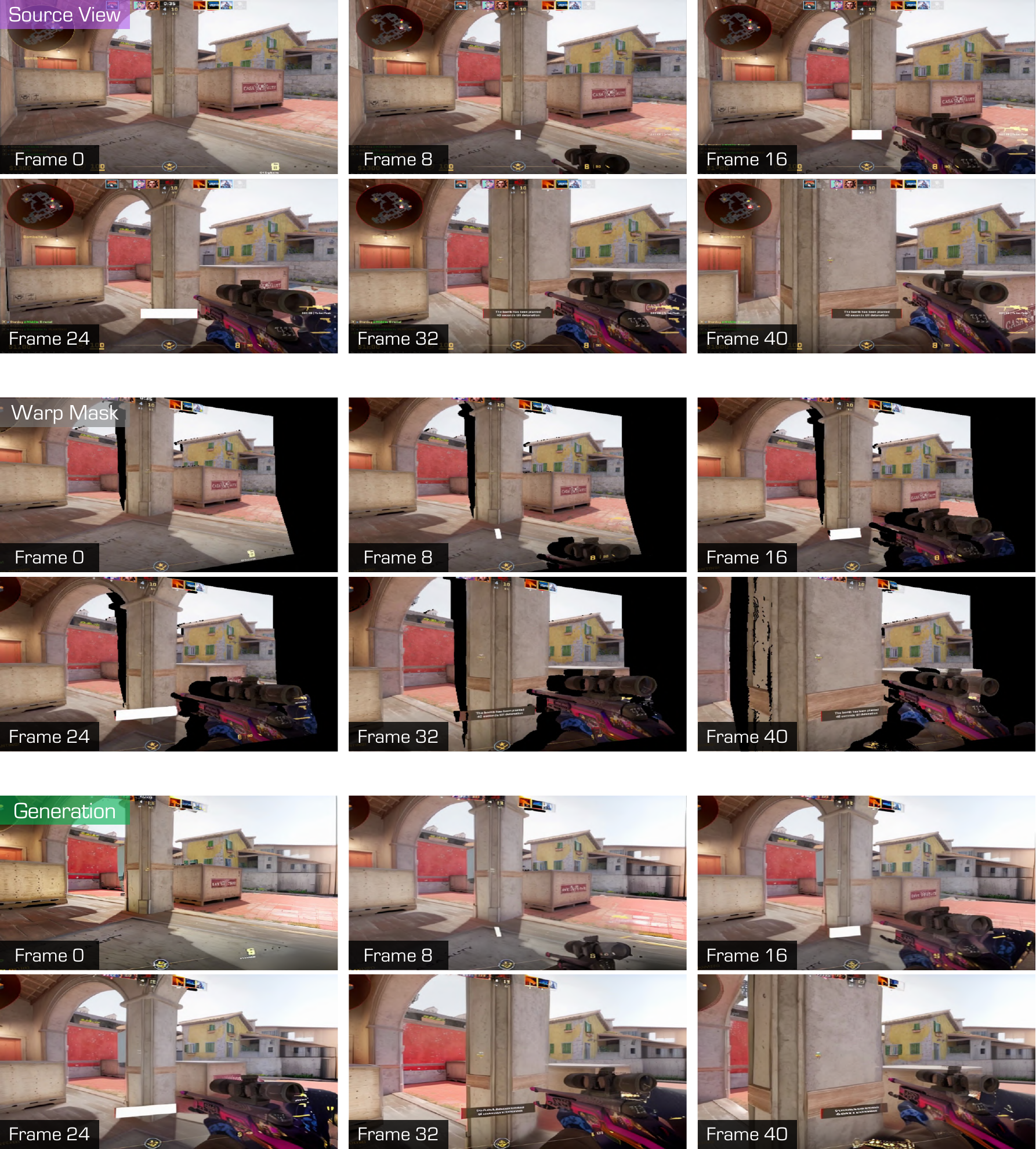}
    \caption{Application use cases of \ours~on computer game generation. The source video is from YouTube.}
    \label{fig:qua_game}
\end{figure*}

\clearpage\clearpage
\begin{figure*}[t]
\vspace{0.16cm}
    \centering
    \includegraphics[width=\linewidth]{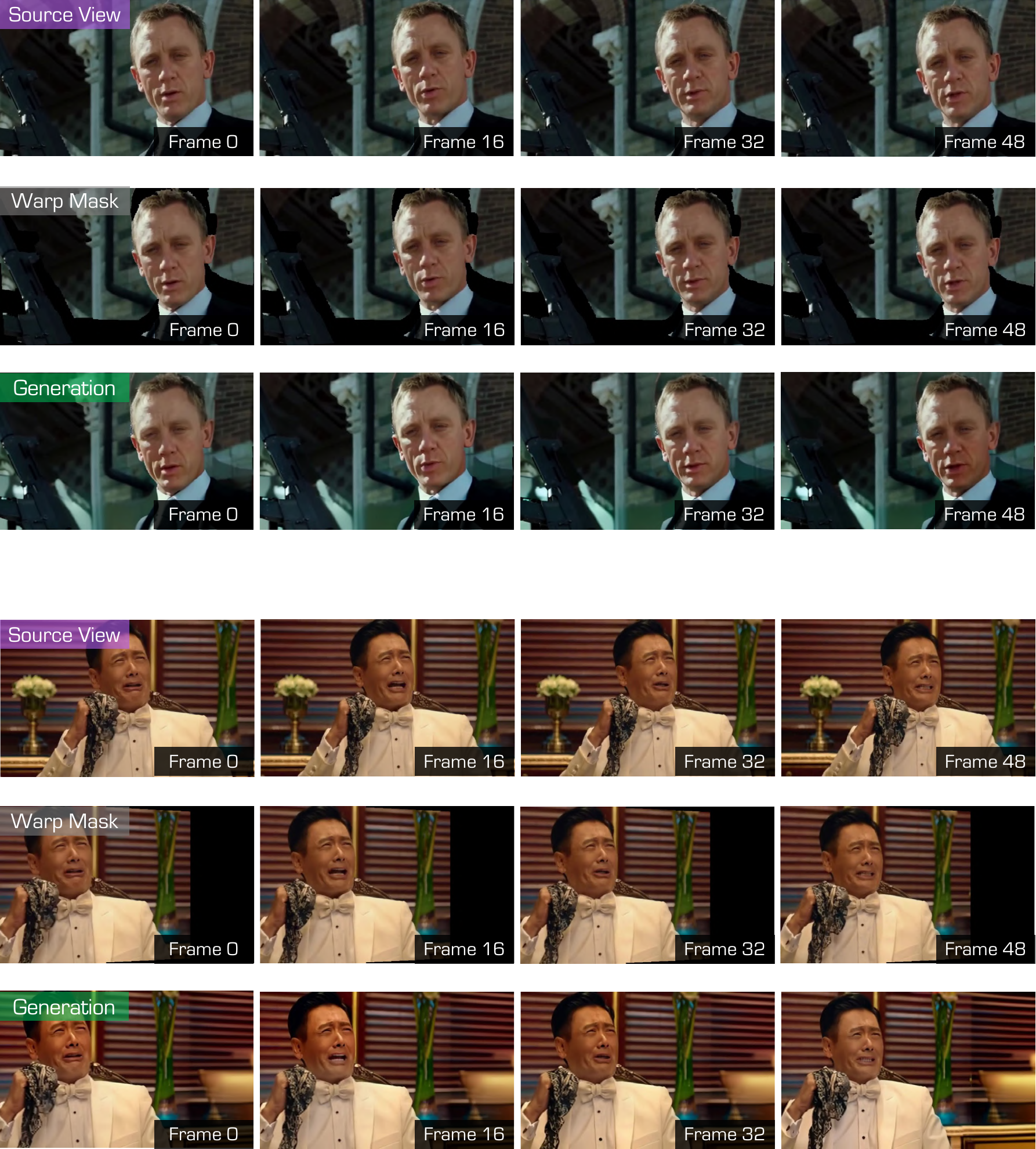}
    \caption{Application use cases of \ours~on movie clip generation. The source videos are from ReCamMaster.}
    \label{fig:qua_movie}
\end{figure*}

\clearpage\clearpage
\begin{figure*}[t]
\vspace{0.16cm}
    \centering
    \includegraphics[width=\linewidth]{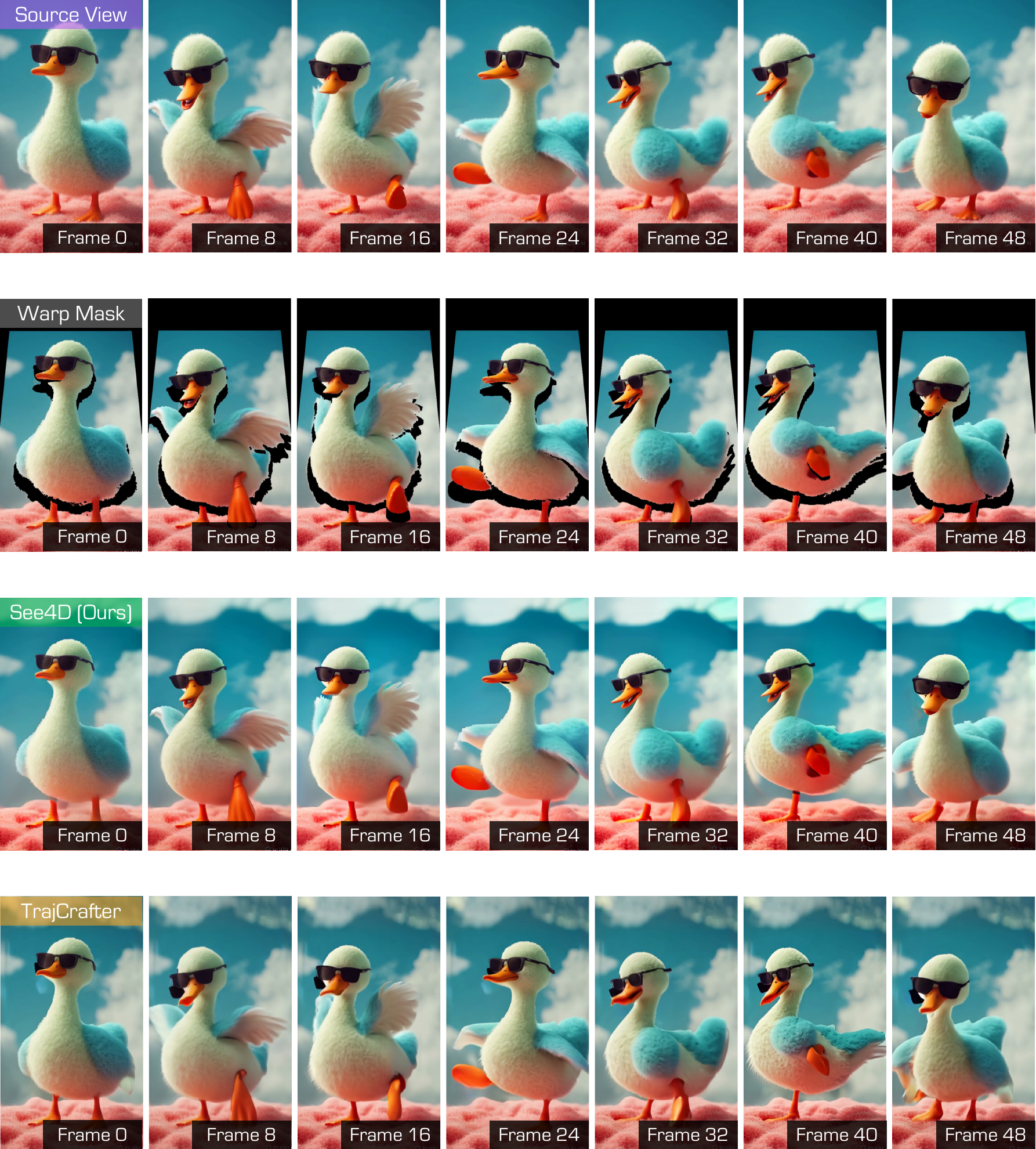}
    \caption{Qualitative comparisons of \ours~and TrajectoryCrafter on video generation. The source video is from KLing.}
    \label{fig:qua_example_1}
\end{figure*}

\clearpage\clearpage
\begin{figure*}[t]
\vspace{0.16cm}
    \centering
    \includegraphics[width=\linewidth]{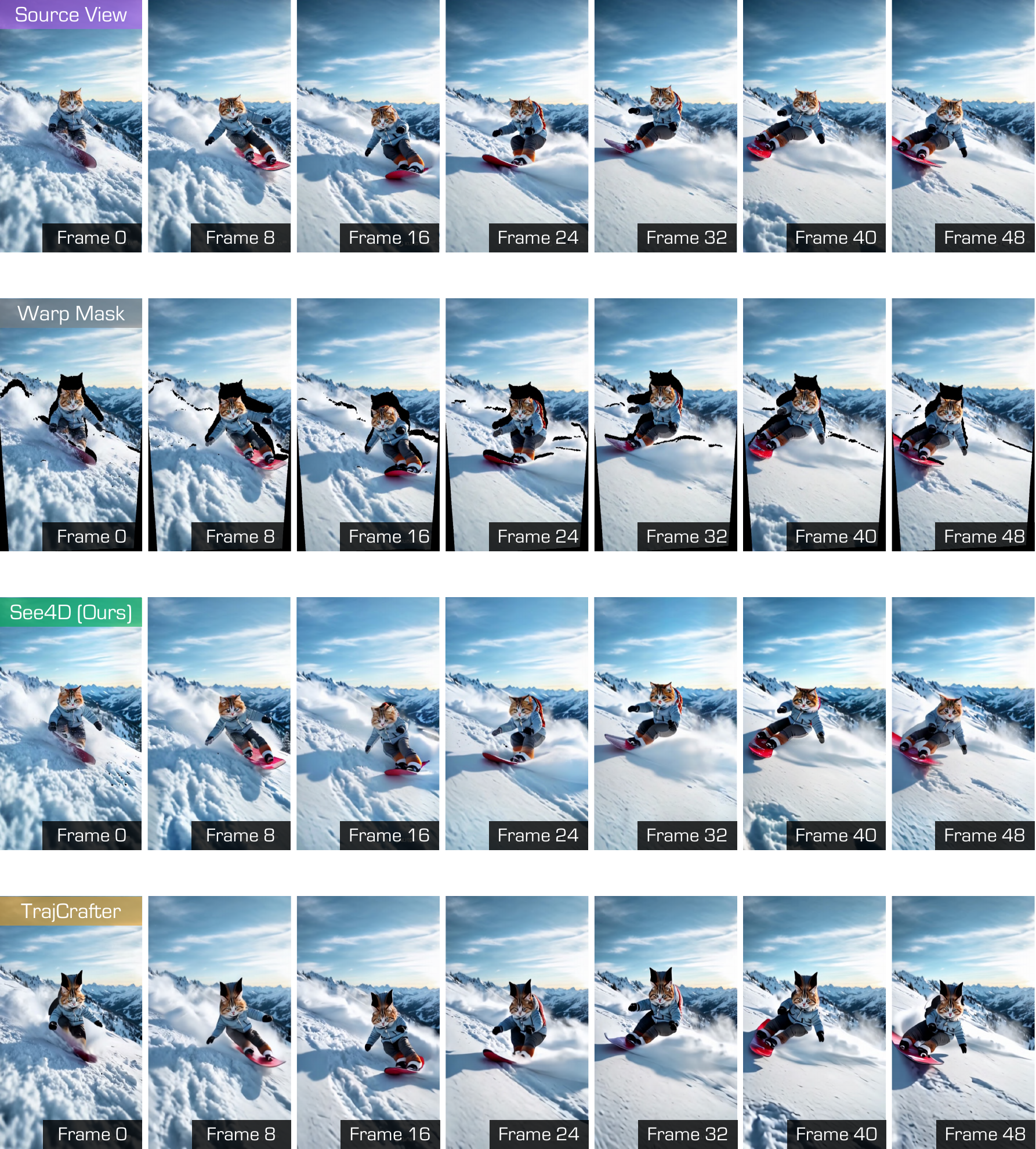}
    \caption{Qualitative comparisons of \ours~and TrajectoryCrafter on video generation. The source video is from KLing.}
    \label{fig:qua_example_2}
\end{figure*}

\clearpage\clearpage
\begin{figure*}[t]
\vspace{0.16cm}
    \centering
    \includegraphics[width=\linewidth]{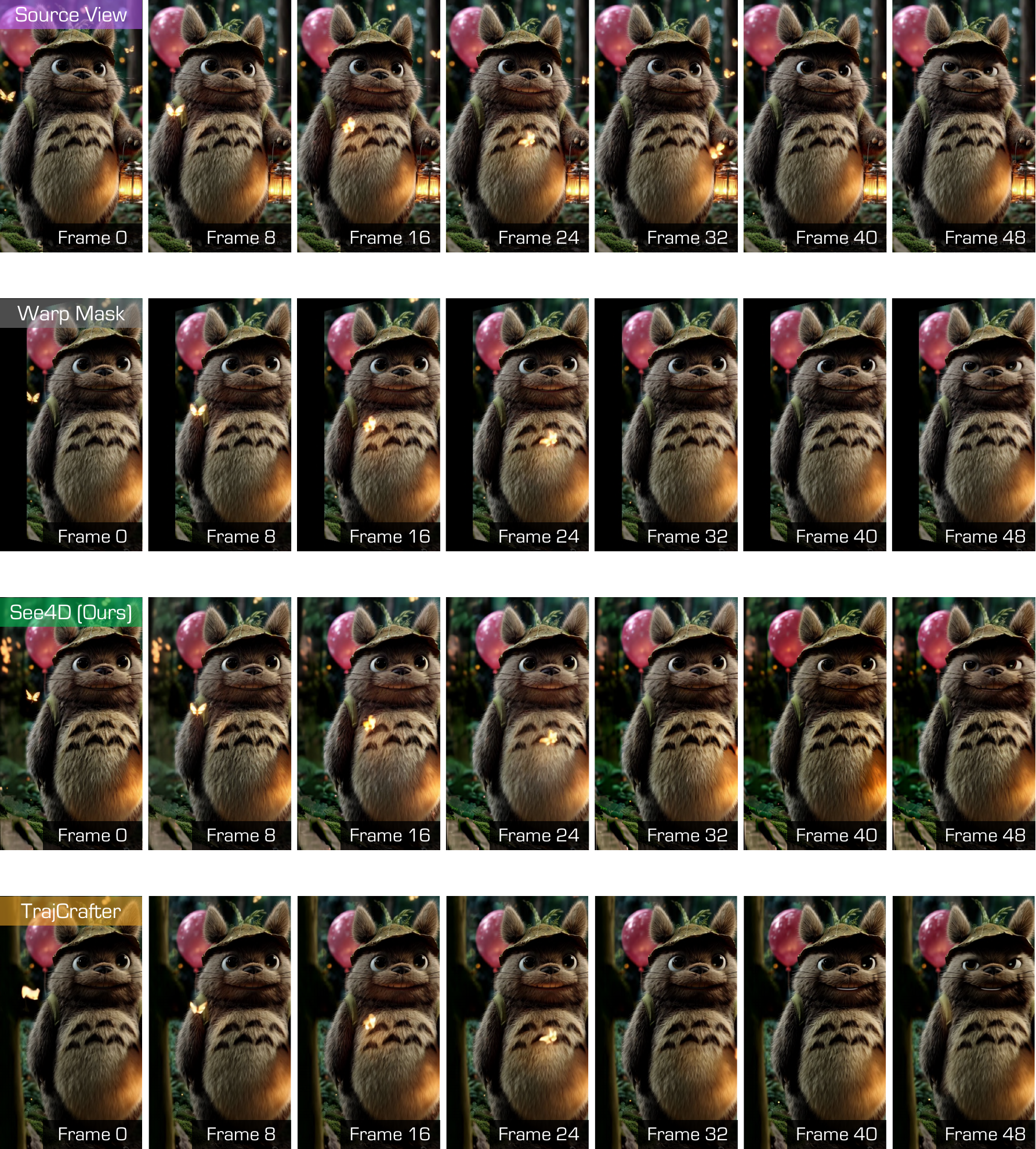}
    \caption{Qualitative comparisons of \ours~and TrajectoryCrafter on video generation. The source video is from KLing.}
    \label{fig:qua_example_3}
\end{figure*}

\clearpage\clearpage
\begin{figure*}[t]
\vspace{0.16cm}
    \centering
    \includegraphics[width=\linewidth]{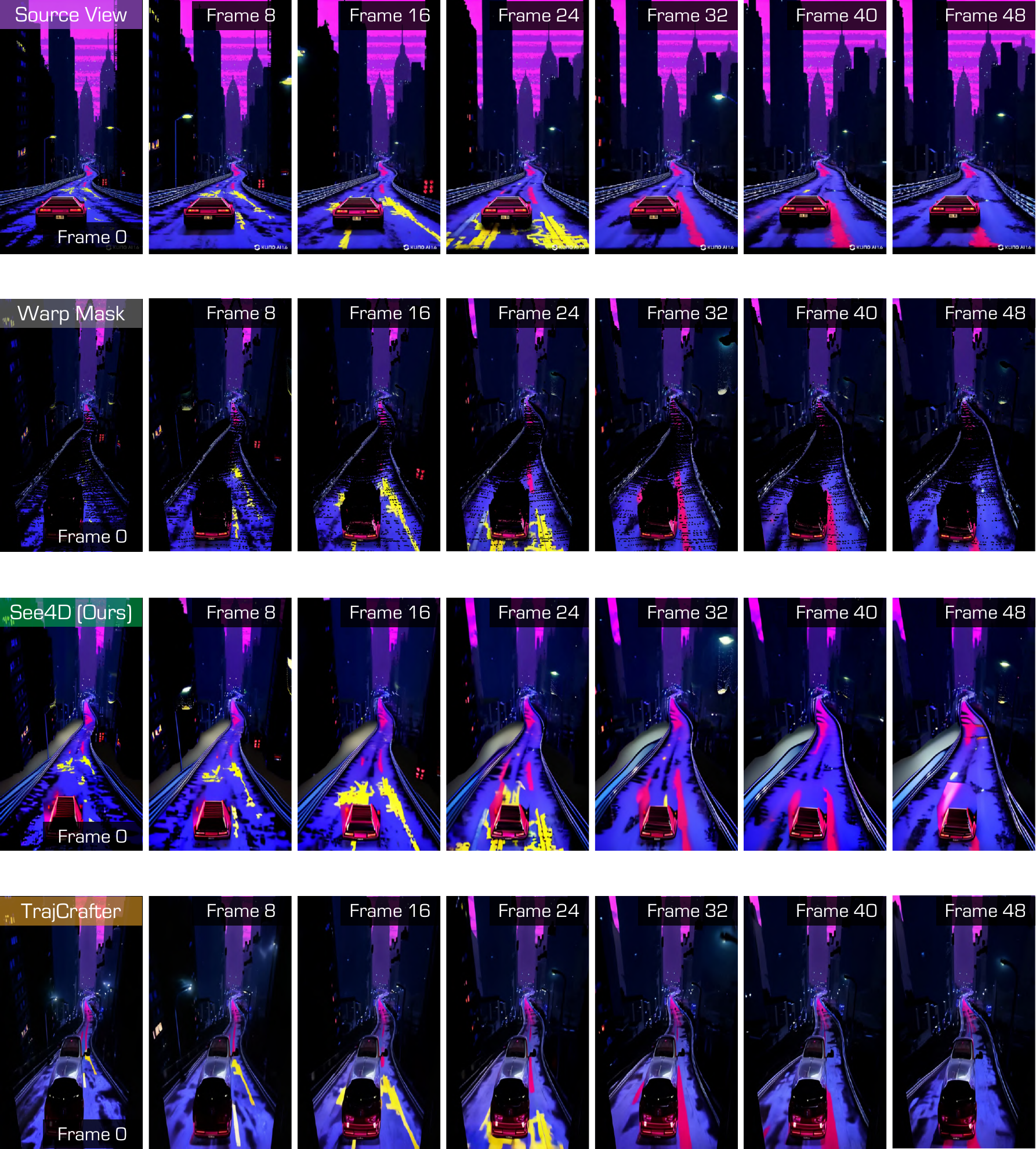}
    \caption{Qualitative comparisons of \ours~and TrajectoryCrafter on video generation. The source video is from KLing.}
    \label{fig:qua_example_4}
\end{figure*}

\clearpage

\printbibliography                

\end{document}